%% file: main-3961-Lachmy.tex
\pdfoutput=1

\documentclass[11pt,a4paper]{article}
\usepackage{times,latexsym}
\usepackage{url}
\usepackage[T1]{fontenc}

    \usepackage[acceptedWithA]{tacl2021v1}
\usepackage{tacl2021v1}

\usepackage{xspace,mfirstuc,tabulary}

\newif\iftaclinstructions
\taclinstructionsfalse 
\iftaclinstructions
\renewcommand{\confidential}{}
\renewcommand{\anonsubtext}{(No author info supplied here, for consistency with
TACL-submission anonymization requirements)}
\newcommand{\instr}
\fi

\iftaclpubformat 

\else

\fi

\usepackage{graphicx}
\usepackage{multirow}
\usepackage{arydshln}
\usepackage{soul} 
\usepackage{amsmath} 
\usepackage{caption}
\usepackage{subcaption}
\usepackage{enotez}

\title{Draw Me a Flower: \\Processing and Grounding Abstraction in Natural Language}

\author{Royi Lachmy\textsuperscript{1,2} \,
 Valentina Pyatkin\textsuperscript{1,2} \,
 Avshalom Manevich\textsuperscript{1} \,
 Reut Tsarfaty\textsuperscript{1,2} \,\\
\textsuperscript{1}Bar-Ilan University, Ramat Gan, Israel \\
\textsuperscript{2}Allen Institute for Artificial Intelligence, Tel Aviv, Israel \\
    {\tt \{royi.lachmy,reut.tsarfaty\}@biu.ac.il}\\
  {\tt  \{valpyatkin, avshalomman\}@gmail.com}\\
  }

\date{}

\begin{document}
\maketitle

\begin{abstract}
Abstraction is a core tenet of human cognition and communication. 
When composing
natural language  instructions, 
humans naturally evoke abstraction to convey complex procedures in an efficient and concise way.
Yet, interpreting and grounding abstraction expressed in NL has not  yet been systematically studied in NLP, with no accepted benchmarks specifically eliciting abstraction in NL.
In this work, we set the foundation for a systematic study of processing and grounding abstraction in NLP. First,
we deliver a novel abstraction elicitation method and present  \textsc{Hexagons}, a 2D instruction-following game.
Using \textsc{Hexagons} we collected
over 4k naturally-occurring  visually-grounded instructions rich with diverse types of abstractions.
From these data, we derive an \textit{instruction-to-execution} task and assess different types of neural models.
Our results show that contemporary models and modeling practices are substantially inferior to human performance, and that models’ performance is inversely correlated  with  the level of abstraction, showing less satisfying performance on higher levels of abstraction. 
These findings are consistent across models and setups, confirming that abstraction is   a challenging phenomenon  deserving further attention and study  in  NLP/AI research.
\end{abstract}

\section{Introduction} 
\label{sec:intro}

\begin{figure}[h!]
\small
  \centering
  \begin{tabular}{ m{3.7cm}  c }
  \hline
      \begin{enumerate}\item[1.]
      {\footnotesize Make a  {\em red flower}, by coloring in red {\em all tiles adjacent} to the 2nd tile from the top in the 2nd column from the left.}
    \end{enumerate}
    &
     \raisebox{-0.45\totalheight}{\includegraphics[width=3.2cm]{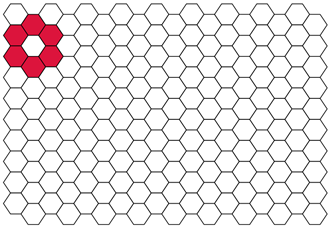}}
    \\ 
      \begin{enumerate}\item[2.]
     {\footnotesize {\em Repeat} this {\em flower} pattern  {\em across the board} to the right, {\em alternating} yellow and red, leaving a blank column {\em between every 2} flowers.}
      \end{enumerate}
    &
         \raisebox{-0.4\totalheight}{\includegraphics[width=3.2cm]{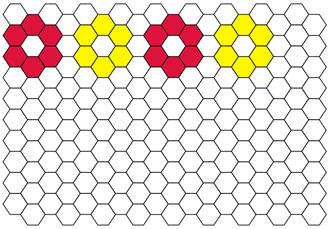}}
    \\
       \begin{enumerate}
       \item[3.]
      {\footnotesize {\em Repeat} this {\em row of flowers}  2 more times, but {\em reverse} the colors {\em in each new row}. You should get 6 red flowers and 6 yellow flowers {\em in total}.}
      \end{enumerate}
    &
         \raisebox{-0.3\totalheight}{\includegraphics[width=3.2cm]{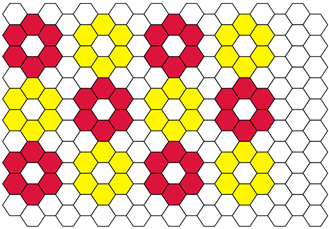}}
    \\ 
     \begin{enumerate}
       \item[4.]
      {\footnotesize Paint the {\em centers} of the flowers using {\em green for red flowers} and {\em blue for yellow flowers.}}
      \end{enumerate}
    &
         \raisebox{-0.55\totalheight}{\includegraphics[width=3.2cm]{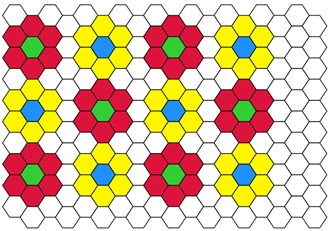}}
    \\  & \\
\hline
  \end{tabular}
  \caption {Abstraction  in the  \textsc{Hexagons} Game. On the left are  instructions for drawing the target image (bottom-right), paired with their grounding on the \textsc{Hexagons} board on the right. {\em Italics} mark expressions of { abstraction}  as referring to, e.g.,   objects (a flower), iterations, and conditions. 
\vspace{-0.2in} 
}
\label{fig:first_page_figure}
\end{figure}

As human-computer interaction in natural language (NL) becomes more and more pervasive,  e.g., via smart devices and chatbots, a cognitive phenomenon known as {\em abstraction}, which is prevalent in human communication and cognition, is taking a  central role in the way users communicate their intentions and needs to artificial agents.  

When communicating in NL, with a human or an artificial agent, a human may issue a request for a single action such as ``send an email'' or ``set my alarm''. However, when  engaging with more {complex} tasks that require multiple actions, humans often evoke \textit{abstraction} in order to communicate their intentions in an economic-yet-precise way.   
Examples for evoking abstraction when issuing  a complex request that consists of multiple actions may be:
``Schedule a group meeting every other Wednesday until the end of the year, unless there are holidays.''. 
For  an autonomous car, an envisioned request might be ``Circle the block looking for a shady parking spot, park at the first spot you see. Try this 4 times, and one more time allowing for non-shady parking. In no luck, try the adjacent block.''. In fact, even the individual request ``send an email'' is in itself  an abstraction over a sequence of multiple individual actions such as :``open your inbox, click new mail, select a recipient,'' etc.
%

Abstraction is defined by \citet{wing2011@CT}
as ``[letting]  one object stand for many. It is used to capture essential properties common to a set of objects while hiding irrelevant distinctions among them''.
In the  calendar example, a  single utterance references multiple meetings in multiple days. Likewise, in the autonomous car example,  the speaker evokes some sort of control structure in order to iterate a process several times.

Abstraction so construed is both critical and pervasive in NL. Referring to multiple instances at once may be done by means of the shape they   form, via a process that iterates them, or via an action/condition applied to select  or manipulate them. 
To illustrate this, Figure~\ref{fig:first_page_figure} showcases an example of a natural language procedure for drawing a target image (the bottom-right image) on an empty board. The instructions start with the construction of
an object, \textit{a red flower}, covering 6 tiles. Then, the Instructor prescribes multiple flower patterns via  {\em repeat} actions that realize a nested `loop'. Finally, the instructor states the color of the flower centers via a `condition' ({\em green for red, blue for yellow}). This example goes to show both the essence and power of abstraction  in NL; here,  merely four  NL utterances suffice to prescribe a complex image over a 180-tiles board.

Despite its importance and widespread use, detecting and grounding abstraction in NL has not yet been systematically studied in NLP. 
Previous studies on grounding instructions target linguistic phenomena  as anaphora and ellipsis \cite{long2016@scone}, spatial relations \cite{jayannavar2020@minecraftBuilder, bisk2016@blockDataset} and referring expressions \cite{haber2019@photobook} but do not specifically elicit abstraction. In studies on navigation 
\cite{anderson_etal2018@navigationRoom2Room, chevalier2018babyai, misraArtzi2018@navigationLaniChai} 
eliciting abstract statements is also  sparse. 
Instructions often refer to specifics of the environment rather than abstract phenomena.

In this work we aim to add a new  facet to the study of natural language understanding,  that  of interpreting abstraction. We set out to provide a foundation for systematically studying the phenomenon of processing and grounding  diverse levels of abstraction  found in naturally-occurring NL utterances. 
Achieving this goal is far from trivial. 
As standard in NLP, we would first need to establish an appropriate dataset for studying this phenomenon. Specifically, we'd like to collect naturally-occurring data that manifest abstraction. But how can we purposefully request for the presence of abstraction in naturally-occurring data? 

To overcome this challenge, we develop an abstraction elicitation protocol by adopting practices from STEM education, specifically from {\em Computational Thinking} (CT) research \cite{wing2011@CT, grover&pea2013@CT}. 
The idea, in a nutshell, is to develop visual stimuli that evoke, and thus cultivate (and   elicit)  higher-order thinking, which is then narrated in NL.
We implement the proposed protocol in a novel \textsc{Hexagons} game, a situated collaborative game where an Instructor provides instructions that should be grounded and executed in a virtual world \cite{long2016@scone, bisk2016@blockDataset, kim2019@codraw, jayannavar2020@minecraftBuilder}. In contrast to  previous studies, we use practices from CT research to design visual triggers of abstraction. Hence, on the one hand, we implicitly call for the presence of abstraction in the instructions, but on the other hand, we provide naturally-elicited abstract instructions from workers not possessing
 formal knowledge of what abstraction  is.   

Using the \textsc{Hexagons} game and the task stimuli we collected over 4k human instructions manifesting a variety of formal abstractions (objects, control structures and functions) expressed naturally and intuitively in NL and grounded on the \textsc{Hexagons} board. 
To showcase how this data may be used for studying abstraction processing in NL, we derive an {\em instruction-to-execution} task, where the model needs to  ground and execute NL instructions on the \textsc{Hexagons} board.
We propose a na\"{i}ve rule-based baseline as well as two neural modeling alternatives --- one based on classification, one on generation --- and assess their performance on the elicited abstraction data.

Our experiments  show that, while  our models perform  better than the na\"{i}ve rule-based baseline, they are substantially inferior to human performance. 
Moreover, we show that models' performance is inversely correlated with the level of abstraction, that is, the models execute concrete instructions quite well, but perform poorly on higher-level abstractions. This holds across different models, setups, task conditions, amount {and type} of training data, and board contexts.
We further observe that the instruction's history is another important factor in models' performance; the longer the history, the better the performance. 

The contribution of this paper is thus manifold.
First, we bring to the fore of NLP research a critical aspect of human-computer communication, namely, the ability to detect, process and ground abstraction in natural language. 
Next, we devise a novel abstraction elicitation methodology and deliver the \textsc{Hexagons} dataset as a novel benchmark to explore the automatic processing of different levels of abstraction. This dataset may also serve broader communities such as AI researchers, linguists, cognitive psychologists and STEM educators in the study of human  processing of abstraction.
Finally, for the instruction-to-execution task we derive from the \textsc{Hexagons} data, we show experimental evidence that unequivocally confirms that abstract instructions in NL are indeed more challenging for current systems to process, and we expose abstraction as an important and challenging dimension  for further study in NLP.\footnote{The data and models along with the collection infrastructure (\textsc{Hexagons} App, Game and tasks)
are publicly available at: \href{https://OnlpLab.github.io/Hexagons}{https://OnlpLab.github.io/Hexagons}}

\section{The  Challenge: Eliciting and Processing Abstraction in NL}
\label{sec:background}

Abstraction is a cognitive phenomenon   related to diverse human activities such as learning, decision making, and behavior regulation (cf. \citealt {burgoon2013abstractionPsychology}). 
In the context of human-computer interaction, and in general in human problem-solving, abstraction is said to be one of a set of cognitive skills known as {\em Computational Thinking}  (CT) skills, defined by \citet*{cuny&wing2010@demystifyingCT} as ``the thought processes involved in formulating problems and their solutions so that the solutions are represented in a form that can be effectively carried out by an information-processing agent.'' 
Abstraction in this context refers to a process of information reduction \citep{burgoon2013abstractionPsychology}, where multiple instances are conceived as arising from a single object, ``consisting of their shared properties while discarding irrelevant distinctions'' \cite{wing2011@CT}. 
    
Abstraction is considered by many as the most important CT skill, allowing  the human to think in terms of objects and concentrating on their essential features, while ignoring irrelevant details \cite{dijkstra1972@humbleProgrammer, denning1989@computingDiscipline, koppelman2010@teachingAbstraction, wing2011@CT, wing2017@CTinfluence}. 
Thus, abstraction leads to speaker's capacity of being more precise and less error-prone \cite{dijkestra@abstractQuote2, haberman2004@proceduralAbstraction} and to designing more concise, elegant and efficient solutions \cite{Ginat2017@multipleAbstraction1}.

\begin{figure}[t]
    \begin{center}
    \scalebox{0.8}{
    \includegraphics[width=7.8cm]{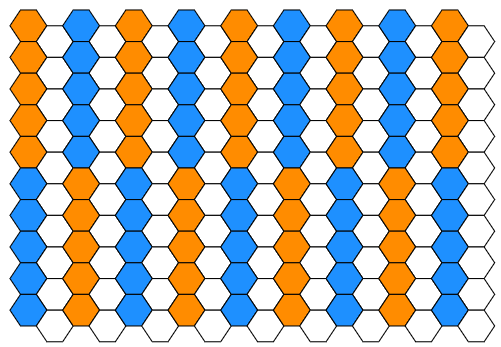}}
    \vspace{0.06in}
    
    \scalebox{0.6}{
    \begin{tabular}{|l|}
    \hline
    \begin{tabular}[c]{@{}l@{}}1.  In the first column paint the first 5 tiles downward orange\\ 2.  In the first column paint the last 5 tiles downward blue\\ 3.  In the 3rd column paint the first 5 tiles blue\\ ...\\ 17.  In the 17th column paint the first 5 tiles orange\\ 18.  In the 17th column paint the last 5 tiles blue\end{tabular}\\ 
    \hline
    \begin{tabular}[c]{@{}l@{}}1.  In first column on left of grid, start with 5 orange hexagons \\ and complete it with 5 blue hexagons. Leave column 2 from \\ left  white.\\ 2.  Repeat these colors and this 5:5 scheme every other column \\ but  alternate the two colors so that each colored column is \\ the opposite  order of the one on either side of it.\end{tabular} \\ 
    \hline
    \end{tabular}
         }
        \caption{Levels of Abstraction in NL. Two drawing procedures for drawing the illustrated image, manifesting low-level and high-level  abstraction.}
        \label{fig:abstraction_example}
    \end{center}
\end{figure}

To illustrate how humans may exhibit different levels of abstraction, consider the simple example in Figure~\ref{fig:abstraction_example}, where a human is requested to describe a pattern on a 2D  \textsc{Hexagons} board.
The first (top) NL procedure expresses low-level abstraction; it refers to each occurrence of a half-column as a unique event. This is a repetitive and lengthy procedure.
In contrast, the second (bottom) NL procedure refers to all the occurrences of this half-column  at once (via `repeat but alternate'), discarding distinctions related to, e.g., tiles'  positions and colors. The result of this abstraction is thus concise, clear and far more efficient.

In a broader sense, in order to express abstraction speakers employ so-called \textit{abstraction mechanisms} --- such as objects, functions and control flow \cite{koppelman2010@teachingAbstraction}  --- and expertise in using them is considered an important part of humans' CT skills \cite{grover&pea2013@CT}. 
Such mechanisms are invoked  in Figure~\ref{fig:first_page_figure}.
And this kind of communication is {not} limited to the simple \textsc{Hexagons} board used in Figures~\ref{fig:first_page_figure}--\ref{fig:abstraction_example}. It is relevant in countless many other domains as  in the aforementioned  calendar and  car examples. 

Due to the prevalence of abstraction in human communication, models would need to process varied levels of abstraction  in order to correctly interpret  NL instructions.
But in order to successfully develop such models, we have to collect data that systematically reflect such phenomena, and to the best of our knowledge, this has not yet been  done in NLP.
The challenge of eliciting abstraction is genuine, as we aspire to elicit {natural language} that reflects {authentic human communication}  of \textit{any} speaker.  
But we cannot simply request crowdworkers to employ {\em abstraction mechanisms} as they are not familiar with these formal concepts. 
On the other hand, explicitly {\em teaching} them to employ abstraction undermines the naturalness of expression.  So, how do we break out of this loop?

\section{The Proposed Methodology}
\label {sec:methodology}

In this work we are interested in creating situations where humans express abstraction in NL spontaneously and naturally, towards learning models that can interpret such abstractions.
To achieve this goal, we turn to  the vast research in human learning and STEM education, on cultivating (and thus eliciting) higher-order CT skills in humans \cite{cuny&wing2010@demystifyingCT, shute2017demystifyingCT}. Eliciting such higher-order thinking requires a careful task design, drawing on literature on the development of instruments that probe and assess humans' CT skills  \cite{Ructtinger2017@taskDesignCT, relkin2019@assessmentCT, Basu2021@taskDesignCT}.

Our designed task stimuli  are carefully crafted to evoke abstraction {\em without} explicitly requesting workers to do so. Our  elicitation methodology extends a recent trend in grounded semantic parsing, where players engage in a referential game ({\em situated collaborative scenarios} in terms of  \citet{jayannavar2020@minecraftBuilder}) where an {\em Instructor} provides instructions that should be grounded and executed in a (simulated) world \cite{long2016@scone, bisk2016@blockDataset, kim2019@codraw, jayannavar2020@minecraftBuilder}. 
The remainder of this section elaborates on the virtual environment we devise, and the task stimuli we design for elicitation.

\subsection{The \textsc{Hexagons} App and Game}
\label{sub-sec:hexagon_game}

In order to collect NL descriptions which express diverse abstraction levels, we design an online  drawing app that enables users to construct increasingly complex images on a \textsc{Hexagons} board, a two-dimensional board paved with hexagonal tiles, of the kind illustrated in Figures~\ref{fig:first_page_figure}--\ref{fig:abstraction_example}.
The \textsc{Hexagons} board contains 18 columns and 10 rows, and the \textsc{Hexagons} App UI provides a drawing interface in which a user may paint tiles using a palette of eight colors.

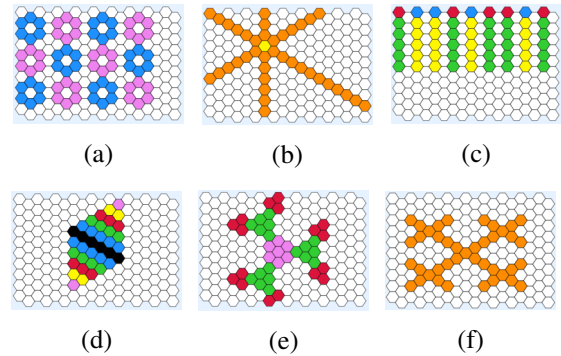
\begin{figure}[t]
    \centering 
    \input{figures/taskgallery}
    \caption{The \textsc{Hexagons} Image Gallery Sample.}
    \label{fig:images_pool01}
\end{figure}

In order to elicit NL instructions, we extended the app with an instruction-following game where a human agent is asked to describe the construction process of a given image (e.g., Figure~\ref{fig:images_pool01}) to a different user of the app, who has access to a similar but blank \textsc{Hexagons} board. The game has two different modes. The first mode is called {\bf Description}, where a user is given an image from a pre-defined pool and has to provide instructions in NL on how to construct the image. Every line break in the textual description initiates a new instruction.
The second mode is called {\bf Execution}, where a user accepts a sequence of instructions one by one, and needs to execute them sequentially to reconstruct the target image on the board.

We refer to each pair of an instruction and a corresponding execution as a {\em drawing step}. We call the sequence of drawing steps composing the full image a {\em drawing procedure}.

\subsection{The  Task Stimuli}
\label{sub-sec:protocol}

The \textsc{Hexagons} app assumes a {single} primitive action that corresponds to the two-place predicate 
{\fontfamily{qcr}\selectfont paint(position, color)}, 
which specifies a color for a specific tile in the 180 hexagon tiles. 
The key idea is to ask Instructors to construct a \textit{complex} image which manifests some \textit{regularity}. The  regularity is intended to encourage Instructors to seek more efficient alternatives to  the primitive-level operations, which then evokes CT skills such as decomposition and abstraction in order to deliver an economic and efficient construction.   %

In what follows we briefly elaborate, for each abstraction mechanism that we target, how we design the form on the {\sc Hexagons} board that potentially evokes this mechanism.

\begin{itemize}
     \item  \textbf{Objects.} 
    Users may refer to  a  set  of  instances at once by means of the form they make (line, circle, triangle, etc.), discarding other details such as the position and color of individual tiles. Objects may be defined in one place, and  referred to elsewhere (as in   Figure~\ref{fig:first_page_figure}).
     
     \item \textbf{Bounded Iterations} (`For' loops). 
     To elicit bounded loops we design images that manifest periodic replication of an object. For example, Figure~\ref{fig:images_pool01}(a) shows a replication of a flower pattern 12 times. 
         
    \item \textbf{Conditional Iterations} (`While' loops).
    To elicit conditional loops we design images that manifest a periodic replication of an object controlled by a certain condition. For example, in Figure~\ref{fig:images_pool01}(b), to replicate the lines with different length one may use the condition `extend the lines out \emph{up to the boundaries of the board}.'
    
    \item \textbf{Conditional Statements} (`if-then'). 
    We design images that manifest random replication of steady variants of an object, where employing a condition  enables users to capture all variants at once. 
    For example, to capture the two  variants of five-tile-long lines in  Figure~\ref{fig:images_pool01}(c), one cannot simply use repetition, as the replication is not periodic. However, noticing that the red and blue `tops' go with the green and yellow `tails' respectively, enables a user to achieve an economic description using a condition on the `top' tiles. 
    
    \item \textbf{Functions.} 
    We design images that manifest replication of objects in different colors or positions, to encourage defining a `block' and then applying it with different parameters. 
    Moreover, we use a particular set of visual functions, of \textit{symmetrical} operations, and particularly reflection and rotation  (e.g., Figures~\ref{fig:images_pool01}(d,e)).
    
    \item {\bf Recursion.} 
    This is a unique type of functions which is challenging to evoke. We approach this challenge by designing three types of stimuli: growing patterns, spirals, and  self-similarity patterns, e.g., fractals (Figure~\ref{fig:images_pool01}(f)).

\end{itemize}

We note that the association between images and targeted  abstraction mechanisms has been defined \emph{a priori}. However, in effect, users may generate instructions with no abstraction or use a different abstraction mechanism to achieve the same result. All in all, since users (and in particular, crowdworkers) aim to be efficient, they tend  (even if not explicitly told) to employ abstractions.

\section{Data Collection and Curation}
\label{sec:collection}

For the data collection we employed English-speaking workers from Amazon Mechanical Turk (MTurk)  and adopted the methodology of controlled crowdsourcing \cite{roit2020@controlledCrowdsource, pyatkin2020@QAdiscourse} to ensure a high quality corpus. 
Specifically, the process includes four stages: {\bf pilot, recruitment,  annotation and consolidation}.

We collect drawing procedures for the task stimuli in a process that comprises of two steps: 
\begin{itemize}
    \item[(1)] 
    In the \emph{Collection} phase
    an Instructor writes instructions for drawing a given image, step by step via the {\bf Description} mode of the game. 
    Following this, the Instructor \textit{aligns} each instruction she has written to its respective execution on the board via the {\bf Execution} mode of the game.
    The result of this process is a {\em drawing procedure} where instructions are coupled with their  actions grounded on the {\sc Hexagons} board.
    \item[(2)]
    In the \emph {Verification} phase
    each drawing procedure from the {\em Collection} phase is given to  {two} Verifiers who do not have access to the original image. The Verifiers are shown the instructions one by one in {\bf Execution} mode. Their task is to execute the instructions step by step  until reconstructing the full image.
    \end{itemize}
    This two-phase process is intended to reveal faulty instructions and to ensure the quality and executability of the collected procedures, by making the Instructor's intentions explicit (step 1), and by exposing disagreements with  Verifiers (step 2).

\begin{figure}[t]
    \centering 
    \input{figures/pool2}
    \caption{A Sample  of Crowdsourced \textsc{Hexagons} Images by MTurk Workers in the Second Round.}
    \label{fig:images_pool02}
\end{figure}
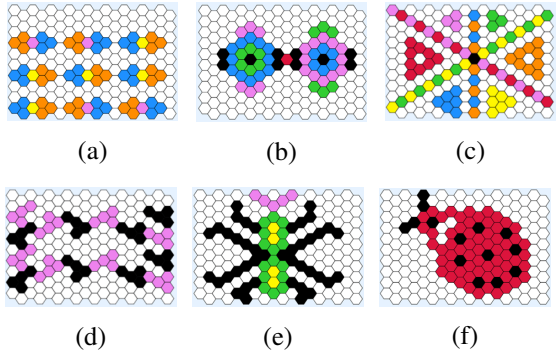

\paragraph{Pilot and Recruitment}
We checked the flow, clarity and feasibility of the data collection in a pilot study, followed by two separate rounds of recruiting Instructors and Verifiers.

To recruit Instructors, we  screened workers by examining understanding and engagement in the Instructor task. 
%
Appropriate candidates had to complete three Instructor tasks, that is, repeat the \textit{Collection} phase with three randomly selected images from our pool. 
In no stage did we formally teach workers what abstraction is.
Instead we engage the workers in several tasks and encourage them to write their instructions \emph{efficiently} and to avoid tiresome repetitions of the  primitive \verb|paint|.  
Out of the 34 candidates, we assembled a group of 28 Instructors exhibiting {\em diverse} levels of abstraction, out of which 24 took an active role during the annotation procedure.

We follow a separate process in recruiting Verifiers using   the \textit{Verification} phase,  instructing candidates to be as accurate as possible while executing  drawing procedures. Out of 27 candidates, we recruited 16 workers 
exhibiting the most precise work.   
The groups of Instructors and Verifiers are disjoint, so drawing procedures are verified by workers other than those who generate them.

\paragraph{Annotation Procedure}
The annotation procedure is based on a {\em Generation-Validation} cycle, which is similar to previous protocols for constructing large-scale corpora by untrained crowdworkers \cite{fitzgerald2018@LargeScaleQASRL}.

Specifically, based on images from our crafted stimuli, drawing procedures are first generated and verified by the Instructors in the \textit{Collection} phase,   and then each procedure is  given to   {two} {\em additional} Verifiers, that work through the \textit{Verification} phase to check the understandability and executability of the  procedures.

The annotation process itself consisted of two rounds.
In the first round we gave each image from our pool to three Instructors in order to generate three different drawing procedures for each image. Each of the generated procedures was verified as usual.
In the second round, we presented Instructors with the opportunity to draw new images on a blank \textsc{Hexagons} board. The goal is to scale-up the extension of the image-pool  with interesting compositions using crowdsourcing. 
Indeed, the collected images in this round reflect similar rationale to our own  set of images (e.g., Figure~\ref{fig:images_pool02}(a-b)) yet demonstrate more complex interactions between structures and patterns (Figure~\ref{fig:images_pool02}(c-d)), with both abstract and figurative images  (Figure~\ref{fig:images_pool02}(e-f)). This new pool of images then passed through the construction and verification phases as usual.

\paragraph{Consolidation}
Having collected the raw dataset, we manually inspected all drawing procedures that had at least one disagreement between an Instructor and {each} of the Verifiers.
Then, we developed a protocol to (i) detect Instructors' errors, (ii) classify the types of errors, and (iii) fix the Instructor execution. 
The protocol was applied to the data by the two first authors. The reported agreement on error classification was 0.95 and 0.98 Cohen's Kappa for the first two tasks and 95\% agreement for the last one. 
Following this protocol, we detected cases where the Instructor's execution is not properly aligned with the instruction, and manually corrected the execution to  match the instruction. 
Types of errors include: Over-/Under-execution where Instructors executes more/less than the instructions require; miscounting of positions on the board; error propagation from previous steps, and others such as using wrong colors. 
All in all we inspected 1461 drawing steps out of which 20.8\% were identified as having Instructors' error and subsequently were manually corrected. 
This process results in data with fully-aligned instruction-execution pairs for each of the instructions in each of the drawing procedures.

\paragraph{Annotation Costs} 
We used Amazon Mechanical Turk to recruit English-speaking workers for this study. 
Participants in the data collection rounds   were paid  higher rates than in  Pilot and Recruitment. The payments for {\em Collection} and {\em Verification} were \$1.5 and \$0.5 respectively.\footnote{We never rejected results or blocked workers. Rather, we used MTurk qualifications as part of our controlled data collection methodology described here.}

\section{The \textsc{Hexagons} Dataset}
\label{sec:dataset}

Our finalized dataset is the collection of all drawing procedures, composed of instructions and their aligned executions collected in both annotation rounds and some from the recruitment stage, after having passed our quality assurance and consolidation process. 
In total, we collected 620  drawing procedures yielding 4177 drawing steps, that is, 4177 instructions with aligned executions, circa 100K tokens.
Table~\ref{table:dataset_volume} 
shows the data statistics.

\begin{table}[t]
    \centering
    \scalebox{0.9}{
    \begin{tabular}{|lcccc|}
    \hline
    & \multicolumn{1}{c}  {Steps}   &  {Procedures} & Tokens & 
    \begin{tabular}[c]{@{}l@{}}Unique\\ Images\end{tabular}
    \\ 
    \hline\hline
    Recruitment  & 1045 & 123  & 21K & 17   \\
    Round I    & 1909   & 304  & 43K  & 101  \\
    Round II   & 1223 & 193    & 36K        & 63 \\
    \hline
    {Total}   & {4177}   & {620}  & {100K} & {175} \\ 
    \hline
    \end{tabular}
}
    \caption{Dataset Statistics}
    \label{table:dataset_volume}
\end{table}

\paragraph{Quantitative Analysis}
In order to quantitatively evaluate the resulting dataset, we define Board-Based and Action-Based metrics that compare two different executions of an instruction.  

Let us define a function \(f(x)\) that accepts a board state \(x\), and translates it into a set of elements  \(\langle position, color\rangle\) that indicate the colored tiles.
Now let \(b\) and \(b'\) be two states of the board, where \(b\) and \(b'\) are considered {\em gold} and {\em hypothesis} respectively. 

\begin{figure}[t]
    \centering
    \scalebox{0.8}{
    \includegraphics[width=7.8cm]{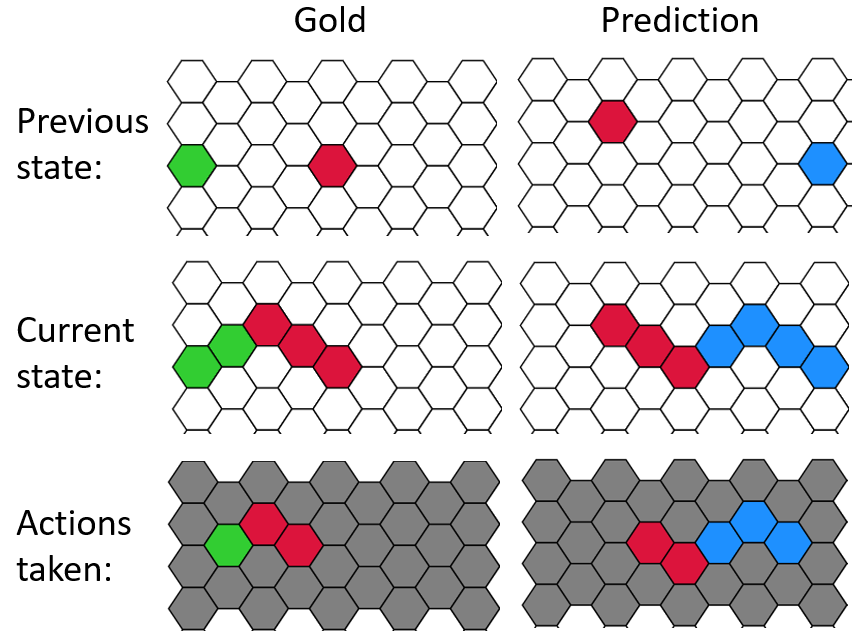}}
    \caption{Board-Based and Action-Based Metrics. In the current state (second row) the intersection between Gold and Prediction is the three red tiles. Therefore, recall and precision are $\frac{3}{5}$, $\frac{3}{7}$ , respectively and Board-Based F1  is $\frac{3}{6}$=0.5. 
    There are three actions taken in Gold and five in Prediction (third row) with one red tile in the intersection of both sets. Thus, recall and precision are $\frac{1}{3}$, $\frac{1}{5}$ , respectively and Action-Based F1 score is $\frac{1}{4}$=0.25. 
    EM for both metrics is 0.}
    \label{fig:metrics}
\end{figure}

Based on the \(f(x)\) function, we can define Precision and Recall as in Equations~\eqref{eq:precision} and \eqref{eq:recall}, respectively, where precision is the percentage of tiles correctly colored  from all those colored in the hypothesis, and recall is the percentage of tiles correctly colored  from all those colored in gold. We then define F1 as usual as the harmonic mean of the two (See illustration in Figure~\ref{fig:metrics}).

\begin{equation}
    \label{eq:precision}\footnotesize{
    Precision(b,b')=\frac{|f(b)\cap f(b')|}{|f(b')|}}
\end{equation}

\begin{equation}
    \label{eq:recall}\footnotesize{
    Recall(b,b')=\frac{|f(b)\cap f(b')|}{|f(b)|}}
\end{equation}
The metrics we report come in two flavours. In Board-Based Metrics, \(f(b)\) picks up all colored tiles in the resulting image after each step.  In Action-Based Metrics, \(f(b)\) focuses {\em only} on the tiles that changed color in the current step, that is, on the instruction's {\em denotation} (rather than the entire board state).

For assessing the quality of the {dataset}, we compare for each drawing step,  the board states (or actions) of the Instructor considered as gold,  to the board states (or actions) of a Verifier, considered the hypothesis. We report the Mean F1 and Exact Match (EM)  averaged  over the entire dataset.
In addition we report  the Max(/Min) Mean F1 and EM which takes into account only the higher(/lower)-scoring Verifier  for each step.

Table~\ref{table:dataset_evaluation} shows the dataset evaluation. The EM metric is more strict than Mean F1. The Max Mean-F1 score of circa 96  indicates that the images can most of the times be reproduced by at least one  human following the instructions, despite the complexity of the instructions. 

\begin{table}[t]
    \begin{center}
    \scalebox{0.8}{
    \input{tables/dataset_evaluation}}
    \caption{Dataset Evaluation.  Min/Max Mean F1 and EM scores are in brackets.}
    \label{table:dataset_evaluation}
    \end{center}
\end{table}

\paragraph{Qualitative Analysis: Overall Phenomena}
\label{sub-sec: qualitative_analysis}

\begin{table}[t]
    \centering
    \scalebox{0.7}{
        \begin{tabular}{|l|l|l|l|l|} \hline
        \multirow{2}{*}{ }General & Type & \# & \# Steps &  \# Drawing  \\ 
        Phenomenon & {} & {} & {} & Procedures \\
        \hline \hline
        \multirow{2}{*} {Abstraction} & Object & 77 & 95 (48.97\%) & 22 (91.67\%)  \\
        {} & Control   & 61 & {}  & {}     \\
        {} & Functions & 49  & {}  & {}    \\ \hline
        \multirow{2}{*}{Goal/Result} & Goal  & 18 & 33 (17.01\%) & 13 (54.17\%) \\
        {} & Result& 17 & {}  & {}  \\\hline
        \multirow{3}{*}{Linguistic}& Anaphora & 63 & 88 (45.36\%) & 15 (62.5\%) \\ 
        {} & Ellipsis & 17 & {}  & {}  \\
        {} & Comparatives & 23 & {}  & {}     \\ 
         \hline
        Spatial & & 225 & 132 (67.69\%) & 23 (95.83\%)   \\ \hline
        \end{tabular}
    }
    \caption{Abstract, Linguistic and Spatial Phenomena in the \textsc{Hexagons} Dataset.}
    \label{table:qualitative_phenomena}
\end{table}

In order to understand the distribution of the elicited NL phenomena in our dataset we sampled  24 drawing procedures with a total of 194 drawing steps (instruction), preserving the internal distribution of stimuli types and annotation rounds. 
We then manually categorized utterances according to abstract, linguistic and spatial phenomena, as summarized in Table~\ref{table:qualitative_phenomena}.

\begin{table}[t]
    \centering
    \scalebox{0.7}{
    \input{tables/examples_abstract_strucutres}
    }
    \caption{Abstraction Mechanisms in  {\sc Hexagons}.}
    \label{table:abstraction_types}
\end{table}

\paragraph{Qualitative Analysis: Levels of Abstraction}
To probe further into the levels of abstraction that are manifested in the dataset, the first two authors annotated all the instructions in the \textit{dev set} 
to one of four levels of abstraction  we identified.

\begin{itemize}
    \item \textit{No abstraction  (0):} In this level Instructors generate concrete instructions which show no abstraction. The instructions specify the coordinates and colors to be painted, in an absolute (e.g., ``Paint the third hexagon in the sixth column green'') or a relative (e.g., ``Paint the tile below this tile red'') fashion.

    \item \textit{Low-Level Abstraction (1)}:
    In this level Instructors refer to a collection of tiles as a single object by means of the topographic shape they form in cases they form vertical or diagonal lines, which are endemic to the \textsc{Hexagons} board (e.g., ``paint the first column from the left green'', ``connect a diagonal line between the two tiles you  just painted'').

     \item \textit{Mid-Level Abstraction (2)}: In this level, Instructors refer to multiple tiles as defining an object (above and beyond lines),  by applying an abstraction mechanisms on multiple tiles  (e.g., first step in Figure~\ref{fig:first_page_figure}).
     
    \item \textit{High-Level Abstraction (3)}: In this level Instructors use diverse abstraction mechanisms   (Table 4) applied to multiple objects which themselves can be complex or abstract (as illustrated in the last three steps in Figure~\ref{fig:first_page_figure}).
\end{itemize}

The annotation to levels was conducted in two stages. In the first stage the dev set was annotated into three levels where the last two levels are combined into a single category (Mid-to-High). In the second stage Mid-to-High cases were split into two levels. The inter-annotator agreement for the first stage is 0.923 Krippendorff’s Alpha \cite{Krippendorff2004@krippendorffAlpha} and for the second stage it is 0.94 Krippendorff’s Alpha. For both coefficient's calculations we use the weighted scheme for ordinal variables \cite{gwet2015@krippendorffAlpha}.

Table~\ref{table:abstraction_level} shows the distribution of abstraction levels we defined within the drawing steps in the dev set. This analysis shows that most of the drawing steps (\(>\) 60\%)
contain abstract instructions, where 50\% contain Mid-to-High abstraction level. 

\begin{table}[t]
    \begin{center}
    \scalebox{0.75}{
     \begin{tabular}{|lccccc|}
    \hline
     & \begin{tabular}[c]{@{}c@{}}No\\ (0)\end{tabular} & 
    \begin{tabular}[c]{@{}c@{}}Low\\ (1)\end{tabular} & 
    \begin{tabular}[c]{@{}c@{}}Mid\\ (2)\end{tabular} &
    \begin{tabular}[c]{@{}c@{}}High\\ (3)\end{tabular} &
    {Total} 
    \\ \hline  \hline

    \hline
    \# Instructions & \begin{tabular}[c]{@{}c@{}}174 \\ (39\%)\end{tabular} & 
    \begin{tabular}[c]{@{}c@{}}47 \\ (10.5\%)\end{tabular} & 
    \begin{tabular}[c]{@{}c@{}}104 \\ (23.3\%)\end{tabular} & 
    \begin{tabular}[c]{@{}c@{}}121\\ (27.1\%)\end{tabular} & 
    \begin{tabular}[c]{@{}c@{}}446\\ (100\%)\end{tabular} \\ 
    \hline
    \end{tabular}
    }
    \caption{Levels of Abstraction in the Dev Set.} 
    
    \label{table:abstraction_level}
\end{center}
\end{table}

\section{Experiments}
\label {sec:experiments}

\paragraph{The Task}
Given the {\sc Hexagons} dataset, we aim to devise models that interpret NL instructions and mimic an Executor's role, in order to assess how standard Pre-trained Language Models (PLM) interpret these utterances. 

To this end, we define a computational task as follows.
Let \(D=d_1 \ldots d_n\) be a sequence of NL instructions of a drawing procedure and let \(b=t_1 \ldots t_{180}\) be a board state naming all tiles' colors on the board at a given state. 
We aim to induce a function \(f(d_1 \ldots d_n)=b_1 \ldots b_n\) that maps a given sequence of instructions to the sequence of board states that indicate the instructions' denotation on the  board.
Since such an \(f\) is overly complex, we model each drawing procedure as a sequence of  {\em instruction-to-execution} steps, where given an instruction \(d_i\) in a procedure, we seek a model for \(g(d_i)=(\{aij\}_{j=1}^{l_i})\) to predict  the  \(l_i\) denoted \textit{actions} \(a_{i1}...a_{i{l_i}}\), where     \(a_{ij}=\langle row_{ij},column_{ij},color_{ij} \rangle\). 
Intuitively, this means that in the execution of each drawing step, we classify each tile to a color label or no\_action.

\paragraph{Input Configurations}
For each of the models, we experiment with several input settings that differ in the type and  extent of  {\em context} provided:
    \textit{(i) No-History}. The input is only the instruction to be executed (current drawing step).
    \textit{(ii) 1-Previous}. The input contains  the instruction to be executed (current step) and one previous instruction. 
    \textit{(iii) Full-History.} The input contains  the instruction to be executed  and all previous instructions in that drawing procedure.   
    \textit{(iv) Oracle Board-state}. The input contains the instruction to be executed   and the gold board-state that is obtained prior to the current drawing step. No previous instruction history is included.
    \textit{(v) Predicted Board-state}. The input contains the instruction to be executed and the predicted board-state,
    predicted  for all  steps so far. No previous instruction history is included.
    \textit{(vi) Full-History + Oracle/Predicted Board-state}, a combination of (iv) and (v) with   full history  (iii).

\paragraph{Data Splits}
We split the dataset into train/dev/test with an 80/10/10 ratio of the drawing procedures (Table~\ref{table:split_statistics}). Following up on recent practices  \cite{finegan-dollak2018@text2SQL, herzig2020span,goldman22lemma} we randomly split the data in a way that {\em avoids} shared stimuli  between the train and the dev/test sets. Specifically, (i) we make sure that there is no image overlap between the three sets (recall that each image is delivered to at least three Instructors; see Section~\ref{sec:collection}), and (ii) we keep the same distribution of images in terms of the abstraction mechanisms they are designed to elicit (Section~\ref{sec:methodology}) as well as the same proportion between the images collected in different annotation rounds.

\subsection{Models}
\label{sub-sec:models}

We  design two  neural models based on two types of  {PLMs}, a BERT-style encoder-only model and a generative encoder-decoder model. For each of these architectures, we modelled the task in a way that is most compatible with it, a classification task for the encoder model (DeBERTa) and  a generation task with the encoder-decoder model (T5). We describe here our two architectures in turn.\footnote{
We fine-tune all models for 20 epochs with early stopping and a batch size of 4.
During inference, when performing conditional sequence generation (with T5) we use beam search with a beam size of 3, without sampling.}

\paragraph{Classification-Based:}
For the classification model we fine-tune DeBERTa\footnote{Specifically we use DebertaForSequenceClassification from the Huggingface Transformers library \cite{wolf2020transformers}, with the \textit{microsoft/mdeberta-v3-base} model.} \cite{he2020deberta} with a classification head to predict an action/no\_action for each of the tiles, resulting in 180 prediction steps for each instruction. The output of each prediction is one  of 9 classes (8 colors and 1 no\_action). 
We define the inputs for the task as follows: The current instruction is prepended with a given tile's coordinates, to indicate for which of the tiles the model is making a prediction; e.g. \textit{<row number> <column number> <current instruction>}. In the \textit{Full-History} setting, the previous instructions are concatenated with a delimiter. When adding the \textit{board-state}, we represent it as a sequence of 180 colors and we additionally mark the given tile with delimiters (e.g.\ \textit{..blue, white, TARGET\_S, red, TARGET\_E}, red..). 

\paragraph{Generation-Based:}
T5 \cite{2020t5} is a generative transformer architecture which uses a text-to-text framework for a variety of NLP tasks.\footnote{We use the T5 V1.1  Model Adapted checkpoint, which we found in preliminary experiments to perform better for our task: \url{https://huggingface.co/google/t5-base-lm-adapt}}
We formulate our text-to-actions task as a text-to-text task using a straightforward input/output scheme: The task's input is put in a template that consists of a prefix and a suffix (\textit{simplify instructions: <current instruction>. simplified instructions:}) and fed as input to the model.\footnote{{The prefix we use is inspired by T5's pre-training scheme; using a natural language task prefix. This template may be viewed as a ``prompt'' \cite{liuNeubig2021@NNPrompting}, but we did not engage with extensive prompt-engineering. We reserve the  investigation of prompts and templates for future work. 
}} The gold output actions are formatted into text by first transforming them into triplets (\textit{<row number> <column number> <color>}) which we then combine into a longer comma separated string (e.g. \textit{0 4 red, 0 5 blue, 1 0 green}) ordering the actions by row and column.
During inference, we generate the most likely continuation for the input  at hand. We then take the generated sequence and parse it into actions, discarding malformed token sequences.
Due to the generative nature of the process, the model's current prediction is conditioned on all  previously predicted actions for a given step.

\paragraph{Baseline and Skyline} 
Our na\"{i}ve baseline model is a deterministic rule-based model based on pattern-matching, inline with previous work \cite{pisl&marecek2017@BlocksRNN}. 
The model we design detects patterns that reflect the basic predicate {\fontfamily{qcr}\selectfont paint(position, color)}, where the position assumes coordinates on a (top-down, left-right) grid.
For example, given the sentence ``In the first column, color the 2nd tile blue'', this model extracts the action \textit{Paint((2, 1), blue)}.
The na\"{i}ve model refers only to the current instruction.
As a skyline we use humans'  performance on the task, presented in terms of the Action-Based Mean F1/EM for the dev and test sets.

\paragraph{Evaluation Metrics}
To evaluate models' performance, we report  Action-Based Mean F1/EM of the predicted actions compared to gold actions. That is,  the Action-Based F1/EM (Section~\ref{sec:dataset}) are averaged over all instructions in the test set. 


    

\begin{table}[t]
    \begin{center}
    \scalebox{0.9}{
    \begin{tabular}{|l||cc|cc|}
    \hline
      & \multicolumn{1}{l}{\#Proc.} & \multicolumn{1}{l}{\#Steps} & \multicolumn{1}{l}{\begin{tabular}[c]{@{}l@{}}Agreed\\ Procedures \end{tabular}} & \multicolumn{1}{l|}{\begin{tabular}[c]{@{}l@{}}Agreed \\ Steps\end{tabular}} \\ 
      \hline\hline
    Train & 496   & 3278 & 392 (79.51\%)   & 87.19\% \\
    Dev   & 62   & 446  & 49 (79.09\%)  & 85.43\%  \\
    Test  & 62 & 453 & 43 (69.35\%)  & 83.22\%  \\
    \hline
    Total & 620  & 4177  & 484 (78.44\%)  & 86.57\%  \\ \hline
    \end{tabular}
    }
    \caption{
    Data Splits Statistics.
    }
    \label{table:split_statistics}
    \end{center}
\end{table}


\begin{table}[t]
    \begin{center}
    \scalebox{0.9}{
        \begin{tabular}{|l||ll|}
        \hline
        & F1    & EM   \\ \hline\hline
        Baseline & 13.15 & 5.96 \\
        Skyline  & 82.06 & 72.3 \\ \hline
        \end{tabular}
    }
    \caption{Baseline and Skyline on Test. }
    \label{table:naive_skyline}
    \end{center}
\end{table}

\section{Results and Analysis}
\label{sec:results}

Table~\ref{table:naive_skyline} shows results of  the na\"{i}ve baseline and the human skyline performance and Table~\ref{table:results_test} shows the performance of the neural models across the different input configurations, on the test set. 

\begin{table}[t]
    \begin{center}
    \scalebox{0.9}{
    \begin{tabular}{|l||ll|ll|}
    \hline
     & \multicolumn{2}{c}{DeBERTa} & \multicolumn{2}{c|}{T5} \\
    & \multicolumn{1}{c}{F1} & \multicolumn{1}{c}{EM} & \multicolumn{1}{c}{F1} & \multicolumn{1}{c|}{EM} \\ 
    \hline\hline
    No-History           & 36.38  & 21.19   & 36.84   & 21.63  \\
    1-Previous & 37.51  & 20.31   & 38.94   & 23.17  \\
    Full-History         & \textbf{46.15}  & 25.39   & 43.6   & \textbf{26.49}  \\
    Predicted Board      & 40.7  & 24.28   & 40.21   & 26.49  \\ 
    Predicted   + Full  & 40.47  & 21.19   & 38.56   & 23.4  \\ 
    \hline 
    Oracle Board         & 43.52  & 22.74   & 43.31   & 28.03  \\
    Oracle  + Full  & 49.55  & 25.61   & 48.11   & 31.35  \\
             \hline
    \end{tabular}
    }
    \caption{Results of DeBERTa and T5 on  Test.}
    \label{table:results_test}
    \end{center}
\end{table}

The results show that all models perform substantially better than the na\"{i}ve rule-based baseline, where the lowest results  (obtained by the No-History condition) are still  23.23 F1 and 15.23 EM points over this baseline on the test-set. 
At the same time, all models are substantially inferior to human performance, where the best model  performance (Full-History)  is 35.91 F1 and 45.81 EM points below human performance on the test set.

DeBERTa and T5 both show the same trends for the different input configurations, with the generative model (T5) often performing better at EM and the classifier (DeBERTa) having higher F1. 

Our ablated experiments on input configurations  are designed to empirically assess the contribution of two kinds of contexts, textual and board-state contexts.
We observe that textual context (previous instructions) is an important factor in model performance; the longer the context is, the better the model performs. 
Performance is lowest when predicting executions on the \textsc{Hexagons} board with only the current instruction as input (No-History).
Adding more context proves to be beneficial, with the Full-History condition having the best realistic (non-oracle) performance. 
This result corroborates  previous findings in studies  which show that models benefit from  textual history  \citep{haber2019@photobook,xu2022@longTermDialogues}.

A different way of providing context for the execution of an instruction is via the state of previous executions on the board. 
Here, we experiment with either providing an oracle board-state at each step, or iteratively feeding the predicted board-states from the previous step to the current step.
While providing the oracle board-state improves performance upon the No-History condition, our results show that it is not as informative as including the full instruction history. 

A possible reason may be that textual instructions often refer back to previously introduced (or decomposed) objects, while board states do not explicitly name these decomposed concepts.
Adding both the oracle board-state and all previous instructions as input results in the best performance, however this is not a realistic setup. The more realistic context setting is that of a predicted board and all previous instructions, but it performs worse than only providing full history, due to error propagation in the predicted states. 

%

\begin{figure}[t]
    \centering
    \scalebox{0.95}{
    \includegraphics[width=1\linewidth]{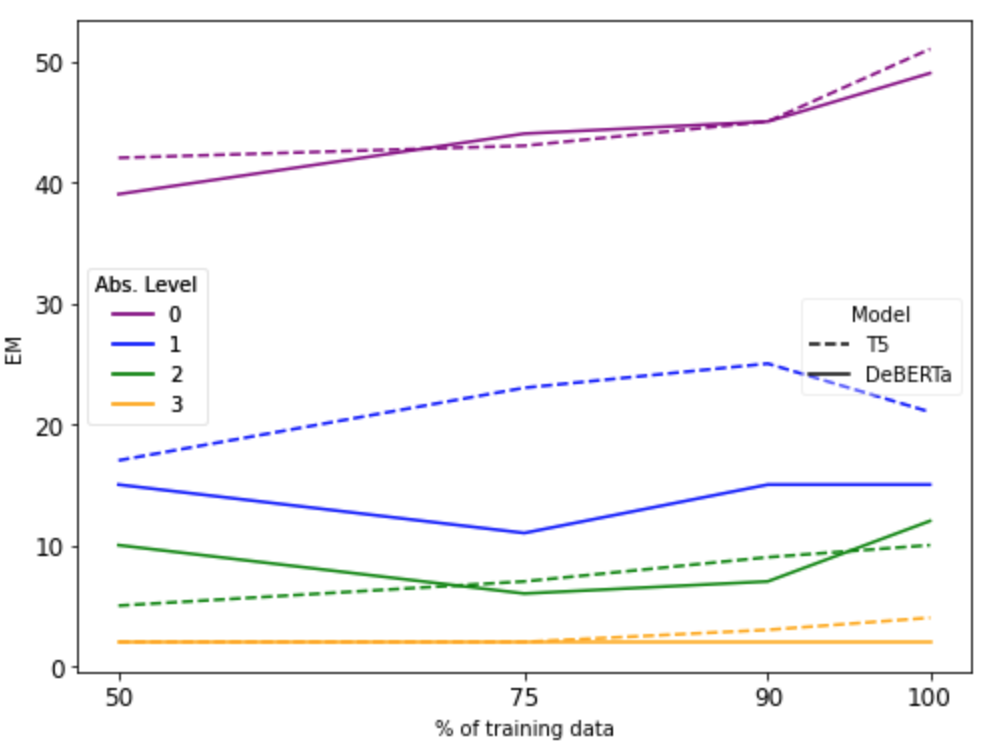}}
    \caption{Qualitative Learning Curve of DeBERTa and T5 on Full History Models, Mean EM.}
    \label{fig:qualitativelc}
\end{figure} 
    \begin{figure}[t]
    \centering
     \scalebox{0.95}{\includegraphics[width=1\linewidth]{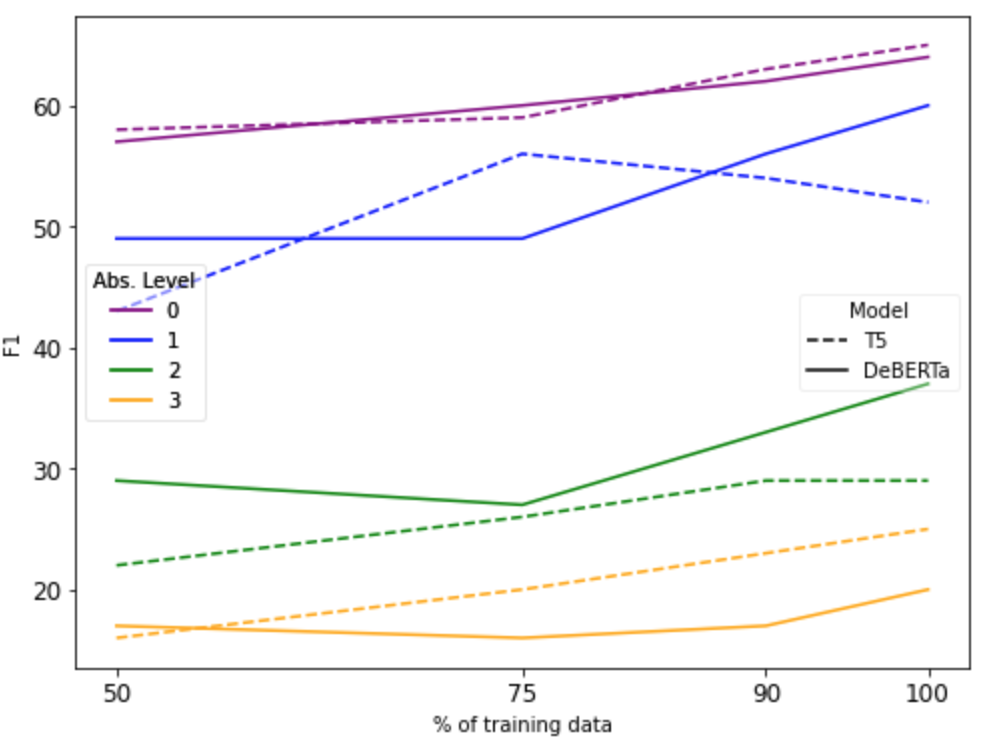}}
    \caption{Qualitative Learning Curve of DeBERTa and T5 on Full-History Models, Mean F1.}
    \label{fig:qualitativelc2}
\end{figure}

\paragraph{Abstraction Levels}
Figures~\ref{fig:qualitativelc}-\ref{fig:qualitativelc2} present the models' performance by abstraction levels (Section~\ref{sec:dataset}) across increasing train set sizes. 
The results on the full train set show that models' performance is {inversely correlated} with the abstraction-level of the instructions; models' performance on executions of concrete primitive-like instructions exceeds those of Mid-to-High level of abstraction. 
This result is significant across models, metrics and input configurations.\footnote{We checked significance by conducting several Kruskal-Wallis tests to compare model performances by abstraction levels for all possible combinations of different models (T5 and DeBERTa), scores (F1 and EM) and input configurations (Full-History and No-History). 
For brevity, Figures~~\ref{fig:qualitativelc}-\ref{fig:qualitativelc2} show only the input configurations of Full-History models.
}

Comparing these results with baseline and skyline by the levels of abstraction (Table~\ref{table:naive_skyline_abstraction_levels}), we observe that models' performances reside between these two boundaries while substantially inferior to skyline across \textit{all} four levels of abstraction. 

\begin{table}[t]
    \begin{center}
    \scalebox{0.9}{
        \begin{tabular}{cccccc}
        \hline
        \multicolumn{1}{|l|}{\multirow{2}{*}{\begin{tabular}[c]{@{}l@{}}Abs.\\ Levels\end{tabular}}} & \multicolumn{5}{c|}{F1}        \\ \cline{2-6}  
        \multicolumn{1}{|l|}{}  & No    & Low   & Mid   & \multicolumn{1}{c|}{High}  & \multicolumn{1}{c|}{All}   \\ \hline  \hline
        \multicolumn{1}{|c|}{Baseline}  & 24.38 & 20.55 & 7.25  & \multicolumn{1}{c|}{3.59}  & \multicolumn{1}{c|}{14.34} \\
        \multicolumn{1}{|c|}{Skyline}  & 87.77 & 93.55 & 85.02 & \multicolumn{1}{c|}{83.52} & \multicolumn{1}{c|}{86.59} \\ \hline
          &       &       &       &  & \multicolumn{1}{l}{}       \\ \hline
        \multicolumn{1}{|c|}{}  & \multicolumn{5}{c|}{EM} \\ \hline  \hline
        \multicolumn{1}{|c|}{Baseline}  & 14.37 & 12.77 & 2.88  & \multicolumn{1}{c|}{0.83}  & \multicolumn{1}{c|}{7.85}  \\
        \multicolumn{1}{|c|}{Skyline}  & 81.32 & 86.17 & 75.96 & \multicolumn{1}{c|}{69.42} & \multicolumn{1}{c|}{77.35} \\ \hline
        \end{tabular}
    }
    \caption{Baseline and Skyline on Dev Sets by Levels of Abstractions.}
    \label{table:naive_skyline_abstraction_levels}
    \end{center}
\end{table}

The results on the gradually increasing train set size show that although in general all levels of abstraction benefit from larger train sets, still model performance on non-abstract instructions is consistently better than instructions exhibiting Mid-to-High-level of abstraction, keeping a mean gap of circa 38 Mean F1. 
Noticeably, the increase for the highest abstraction level is very mild.  especially for EM.  This hints that no substantial learning is happening at the highest abstraction level, and a different architecture or training regime, geared towards abstraction, is needed.

We manually inspected executions of our models with respect to the levels of abstraction of the instruction.
Looking at the successful executions of  high-level instructions, we observe that those instructions are mainly instances where no actions should be performed, e.g., Goal/Result declarations, or instances where very common objects (e.g., flowers) are defined and drawn. More complex functions, such as \textit{repetitions with conditions}, are harder for the models to interpret.

\begin{figure}[t]
    \centering 
    \input{figures/abstractionleap}
    \caption{Processing Different Abstraction Levels. DeBERTa's executions (left) and board-states (right) for different levels of abstraction.
    Instruction  (a)-(b): ``Use green to fill in the 2nd and 3rd spots on the 3rd column, 1st and 2nd spots on the 4th column, and 2nd spot on the 5th column.'', F1=1;  (c)-(d): ``Create the same shape with green on all the purple and orange spots.'', F1=0.08.
    }
    \label{Fig:abstractionleap}
     
    \end{figure}
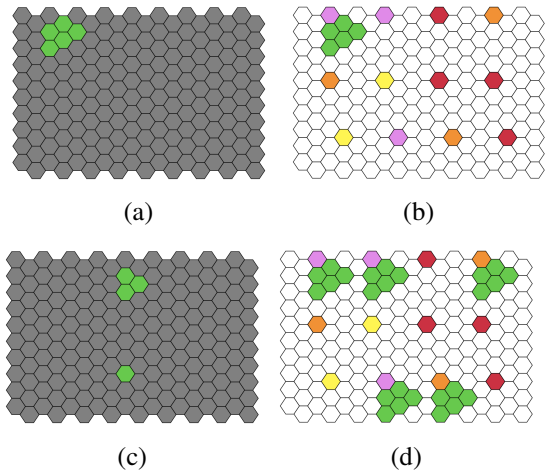

An example of how executions differ between different abstraction levels is displayed in Figure~\ref{Fig:abstractionleap}. The model correctly executes the first instruction which contains no abstraction. The next instruction is of high abstraction 
including replication of the triangle object. 
The model does not manage to identify the spots where to attach the new triangles, or to generate appropriate triangles. 

These  findings are all consistent with the claim  that abstract instructions pose a challenge for current NLP technology, {\em orthogonally} to data size and various other factors.

\section {Related Work}
Studying abstraction in collaborative communication  is related to previous studies on collaborative  games that focus on how interlocutors generate referring expressions accepted and understood by both the speaker and hearer \cite{clark&Wilkes-Gibbs1986@referringCollaborative, khaniGoodmanLiang2018@InfoJigsaw, haber2019@photobook,  Takuma&Akiko2019@oneCommon}. 
In such situations a speaker attempts to generate the shortest refererring  expression that will sufficiently communicate their intention.
%
%
This phenomenon of minimizing speakers' effort    is inline with Grice's (\citeyear{grice1975@logic})
maxim of  \textit{quantity}, stating that speakers will give as much information as  needed and not more.

The settings of collaborative games are very common in creating datasets for grounded semantic parsing as navigation tasks \cite{anderson1991hcrc, MacMahon2006@sail, anderson_etal2018@navigationRoom2Room, chevalier2018babyai, misraArtzi2018@navigationLaniChai,  chen2019@touchdown, tzuf2019@RUN, suhr2019@CerealBar}, the 2-D/3-D blocks world \cite{bisk2016@blockDataset, bisk2016@blockGroundingModel, bisk2018@blockAbstract,jayannavar2020@minecraftBuilder} and other instruction-following scenarios \cite{long2016@scone,kim2019@codraw}. 
Some of these studies observe abstraction as a  phenomenon that indeed occurs in NL instructions \cite[e.g.,][]{anderson_etal2018@navigationRoom2Room}, implying that abstraction is a cross-domain and hence critical phenomenon for natural language understanding.   
However, eliciting naturally-occurring NL instructions that reflect a variety of abstraction levels in a systematic way is novel to the \textsc{Hexagons} data.

To confirm this, we inspected the 2-D Blocks dataset \cite{bisk2016@blockDataset, bisk2016@blockGroundingModel,pisl&marecek2017@BlocksRNN} which most resembles our setting. 
Sampling 594 instructions (5\% of the train set),
we found out that almost all the instructions (96.5\%) map to actions of shifting a single block to some location, with some spatial expressions (e.g., ``place box 17 three spaces above box 20''). 
Such instructions are labeled ``no abstraction'' in our protocol (see Section~\ref{sec:dataset}). Notably, a fraction of the sample does express some low-level (2.7\%) and mid-level (0.8\%) abstraction.

Two other studies that are particularly related to our work are by \citet{wang2017@naturalizingVoxelurn} and \citet{wang2016@SHRDLURNWittgensteinLanguageGame}. In the VoxeLurn study \cite{wang2017@naturalizingVoxelurn}, a community of users is interacting with a computerized agent to deliver constructions in a 3-D blocks world. The community gradually and collaboratively builds increasingly complex and more {\em abstract} language from a core programming language via a process called ``naturalization''. 
SHRDLURN \cite{wang2016@SHRDLURNWittgensteinLanguageGame} exhibits similar constructions but on an individual rather than a community effort. 
Both studies indeed  address abstraction, but from an {\em opposite} direction to ours; while these works assume a strict narrow and synthetic language and build abstractions bottom-up, 
our work aims to tackle the opposite direction, uncovering  abstractions  that are expressed in unrestricted informal NL and grounding them in an executable `backend'. Thus, these studies and ours exhibit orthogonal ways to address abstraction.

\section {Conclusion} 
\label{sec:discussion}

We bring to the fore of NLP  a novel and critical aspect of human-computer communication, namely, the ability to automatically detect, interpret and ground {abstraction} in NL. 
We devise an abstraction elicitation methodology and deliver a novel benchmark,  \textsc{Hexagons}, manifesting the denotation of instructions rich and diverse in their levels of abstraction.
Our results on the   {\em instruction-to-execution} task derived from these data show that the models' performance is significantly {\em inversely} correlated with the level of abstraction, and this holds across  models, contexts, and data sizes. 
This work opens a manifold of directions for future research such as generating human-like abstractions or detecting the level of abstraction, as well as studying abstraction in adjacent fields as linguistics, cognitive science and NL programming. 

\section*{Acknowledgements}
We thank the audience of the BIU-NLP Seminar, the BIU Linguistics Colloquium, and the TAU-NLP Seminar, for fruitful discussion of this work. 
We specifically thank  Yoav Goldberg for his critical comments on an earlier draft. 
We would also like to thank our action editor, Christopher Potts, and the anonymous reviewers for their invaluable suggestions and feedback. 
This research is funded by the the European Research Council, ERC-StG Grant no.\ 677352, for which we are grateful.

\bibliography{tacl2021}
\bibliographystyle{acl_natbib}
\end{document}

%% file: figures/taskgallery.tex
\begin{subfigure}{0.14\textwidth}
  \includegraphics[width=\linewidth]{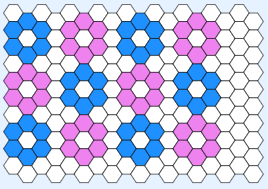}
  \caption{}
\end{subfigure}\hfil 
\begin{subfigure}{0.14\textwidth}
  \includegraphics[width=\linewidth]{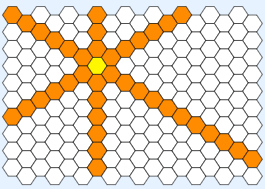}
  \caption{}
\end{subfigure}\hfil
\begin{subfigure}{0.14\textwidth}
  \includegraphics[width=\linewidth]{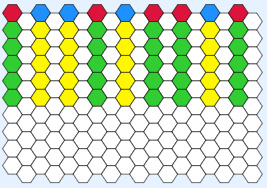}
  \caption{}
\end{subfigure}

\medskip
\begin{subfigure}{0.14\textwidth}
  \includegraphics[width=\linewidth]{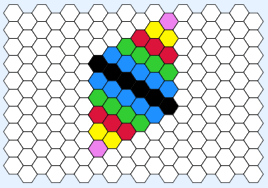}
  \caption{}
\end{subfigure}\hfil 
\begin{subfigure}{0.14\textwidth}
  \includegraphics[width=\linewidth]{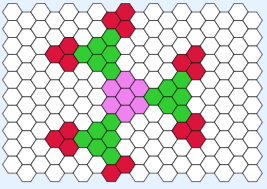}
  \caption{}
\end{subfigure}\hfil
\begin{subfigure}{0.14\textwidth}
  \includegraphics[width=\linewidth]{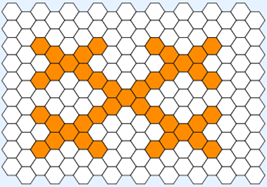}
  \caption{}
\end{subfigure}

%% file: figures/pool2.tex


  



\begin{subfigure}{0.14\textwidth}
  \includegraphics[width=\linewidth]{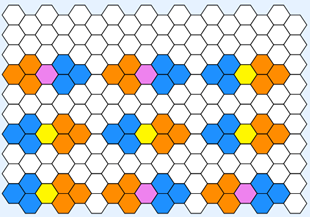}
  \caption{}
\end{subfigure}\hfil 
\begin{subfigure}{0.14\textwidth}
  \includegraphics[width=\linewidth]{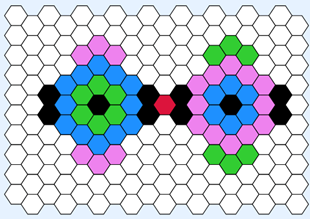}
  \caption{}
\end{subfigure}\hfil
\begin{subfigure}{0.14\textwidth}
  \includegraphics[width=\linewidth]{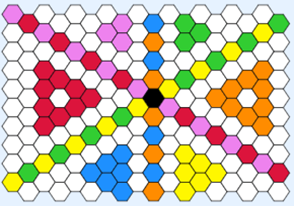}
  \caption{}
\end{subfigure}

\medskip
\begin{subfigure}{0.14\textwidth}
  \includegraphics[width=\linewidth]{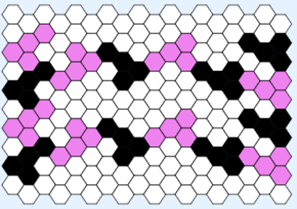}
  \caption{}
\end{subfigure}\hfil 
\begin{subfigure}{0.14\textwidth}
  \includegraphics[width=\linewidth]{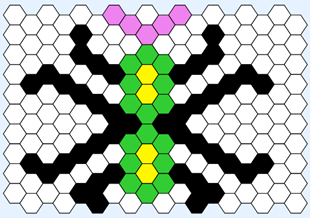}
  \caption{}
\end{subfigure}\hfil
\begin{subfigure}{0.14\textwidth}
  \includegraphics[width=\linewidth]{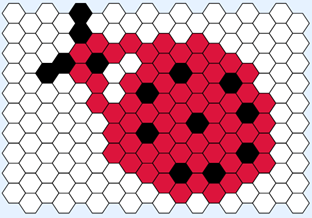}
  \caption{}
\end{subfigure}

%% file: tables/dataset_evaluation.tex
    

\begin{tabular}{| l c c|}
     \hline
      &  Mean F1  & Mean Exact Match \\[0.5ex] 
     \hline\hline
     {Board-Based} & 91.11 [85.85, 96.37]  & 72.32 [58.07, 86.57]\\
    {Actions-Based} &  84.46 [74.71, 94.21]  & 75.98  [62.58, 89.38]\\ 
    \hline
\end{tabular}

%% file: tables/examples_abstract_strucutres.tex
\begin{tabular}{|l||l|}
\hline
 Abs. Mechanisms & Examples  \\ 
\hline \hline
Bounded Iterations & 
    \begin{tabular}[c]{@{}l@{}}``Make four more caterpillars below  \\  the original  leaving an empty space \\ between every two caterpillar.''
    \end{tabular}
    \\
Cond.\ Iterations &
    \begin{tabular}[c]{@{}l@{}}``Repeat this pattern until you \\run out of room on grid.'' \end{tabular}
    \\ 
Cond.\ Statements &
    \begin{tabular}[c]{@{}l@{}}``Directly beneath the painted tiles,   \\  paint green and yellow vertical  columns \\ of five touching tiles, using green below  \\ the red tiles and  yellow below the blue tiles.''
    \end{tabular}
    \\
Objects &
    \begin{tabular}[c]{@{}l@{}}``make a blue dog bone shape''
    \end{tabular}
    \\
Symmetry &
    \begin{tabular}[c]{@{}l@{}}``Reflect the multi-colored triangle you  \\ made in the previous two  steps   \\ symmetrically over the black diagonal.''
    \end{tabular}
    \\
Recursion &
    \begin{tabular}[c]{@{}l@{}}``Form an X by ... At each corner, \\ ... form 4 smaller X shapes''
    \end{tabular}
    \\ 
    \hline
    \end{tabular}

%% file: figures/abstractionleap.tex
\begin{subfigure}{0.21\textwidth}
  \includegraphics[width=\linewidth]{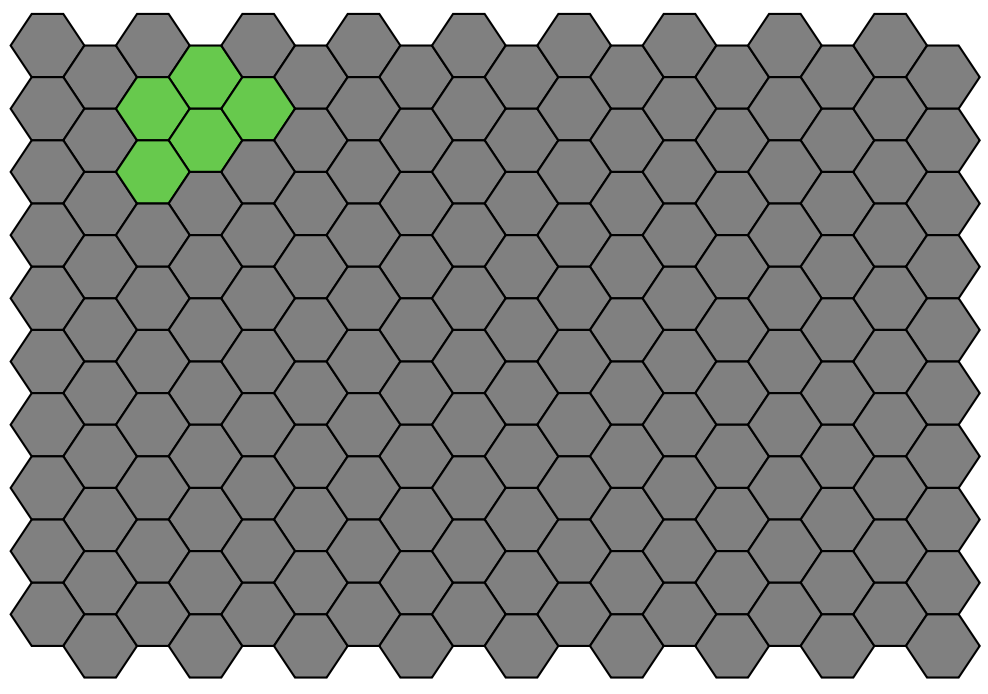}
  \caption{}
\end{subfigure}\hfil 
\begin{subfigure}{0.21\textwidth}
  \includegraphics[width=\linewidth]{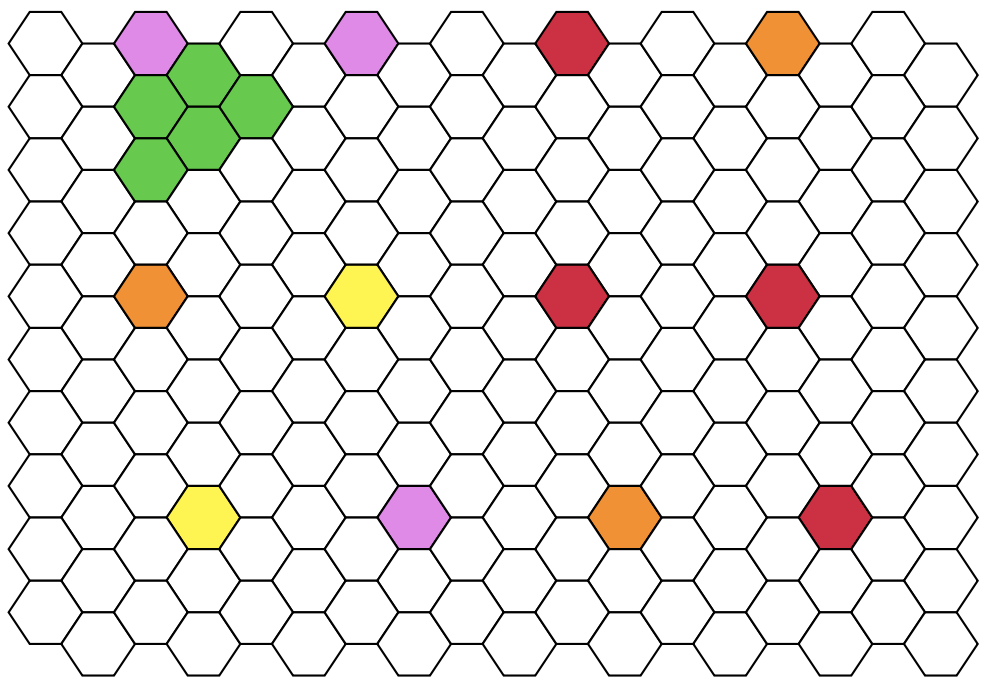}
  \caption{}
\end{subfigure}\hfil

\medskip
\begin{subfigure}{0.21\textwidth}
  \includegraphics[width=\linewidth]{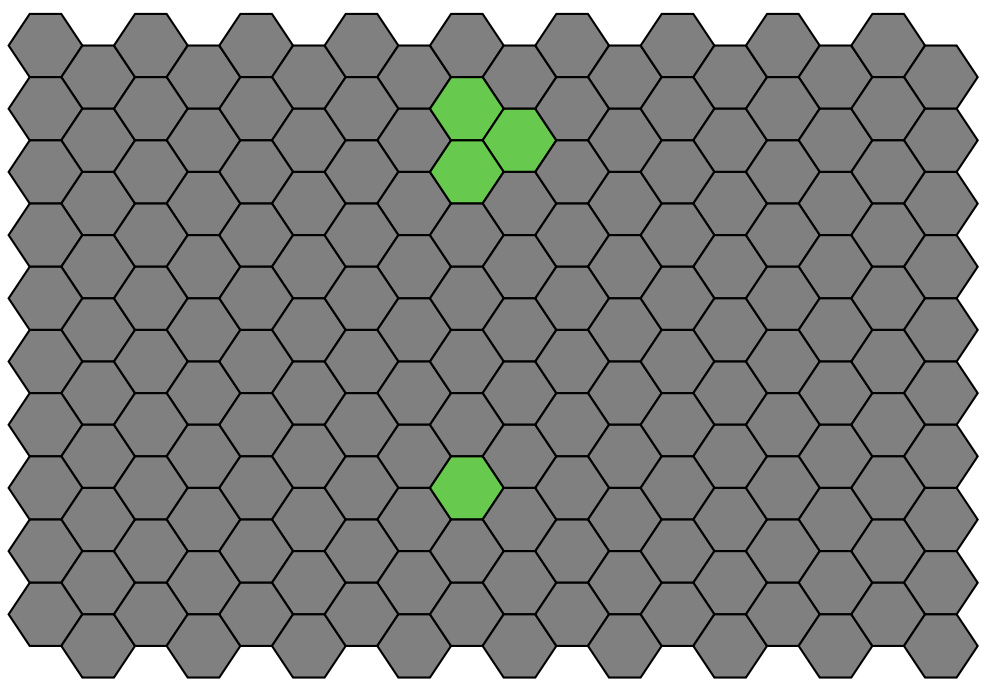}
  \caption{}
\end{subfigure}\hfil 
\begin{subfigure}{0.21\textwidth}
  \includegraphics[width=\linewidth]{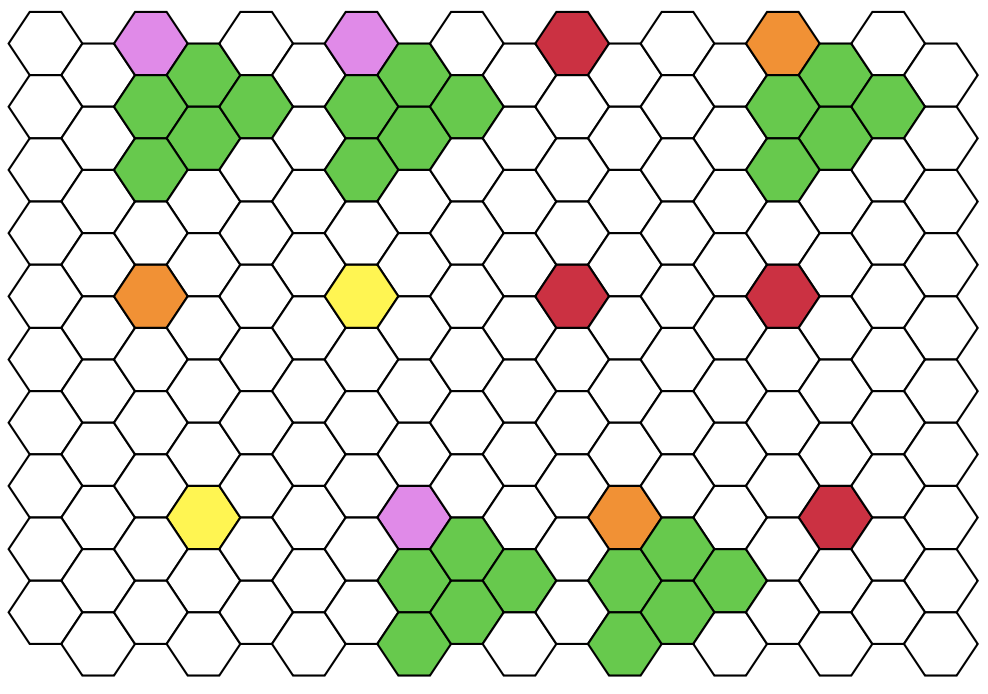}
  \caption{}
\end{subfigure}\hfil

%% file: main-3961-Lachmy.bbl
\begin{thebibliography}{50}
\expandafter\ifx\csname natexlab\endcsname\relax\def\natexlab#1{#1}\fi

\bibitem[{Anderson et~al.(1991)Anderson, Bader, Bard, Boyle, Doherty, Garrod,
  Isard, Kowtko, McAllister, Miller, Sotillo, Thompson, and
  Weinert}]{anderson1991hcrc}
Anne~H. Anderson, Miles Bader, Ellen~Gurman Bard, Elizabeth Boyle, Gwyneth
  Doherty, Simon Garrod, Stephen Isard, Jacqueline Kowtko, Jan McAllister, Jim
  Miller, Catherine Sotillo, Henry~S. Thompson, and Regina Weinert. 1991.
\newblock \href {https://doi.org/10.1177/002383099103400404} {The {HCRC} map
  task corpus}.
\newblock \emph{Language and Speech}, 34(4):351--366.

\bibitem[{Anderson et~al.(2018)Anderson, Wu, Teney, Bruce, Johnson,
  S{\"u}nderhauf, Reid, Gould, and van~den
  Hengel}]{anderson_etal2018@navigationRoom2Room}
Peter Anderson, Qi~Wu, Damien Teney, Jake Bruce, Mark Johnson, Niko
  S{\"u}nderhauf, Ian~D. Reid, Stephen Gould, and Anton van~den Hengel. 2018.
\newblock Vision-and-language navigation: {I}nterpreting visually-grounded
  navigation instructions in real environments.
\newblock \emph{2018 IEEE/CVF Conference on Computer Vision and Pattern
  Recognition}, pages 3674--3683.

\bibitem[{Basu et~al.(2021)Basu, Rutstein, Xu, Wang, and
  Shear}]{Basu2021@taskDesignCT}
Satabdi Basu, Daisy~W. Rutstein, Yuning Xu, Haiwen Wang, and Linda Shear. 2021.
\newblock A principled approach to designing computational thinking concepts
  and practices assessments for upper elementary grades.
\newblock \emph{Computer Science Education}, 31(2):169--198.

\bibitem[{Bisk et~al.(2016{\natexlab{a}})Bisk, Marcu, and
  Wong}]{bisk2016@blockDataset}
Yonatan Bisk, Daniel Marcu, and William Wong. 2016{\natexlab{a}}.
\newblock \href {http://www.aaai.org/ocs/index.php/WS/AAAIW16/paper/view/12652}
  {Towards a dataset for human computer communication via grounded language
  acquisition}.
\newblock In \emph{Symbiotic Cognitive Systems, Papers from the 2016 {AAAI}
  Workshop, Phoenix, Arizona, USA, February 13, 2016}, volume {WS-16-14} of
  \emph{{AAAI} Workshops}. {AAAI} Press.

\bibitem[{Bisk et~al.(2018)Bisk, Shih, Choi, and
  Marcu}]{bisk2018@blockAbstract}
Yonatan Bisk, Kevin~J. Shih, Yejin Choi, and Daniel Marcu. 2018.
\newblock \href
  {https://www.aaai.org/ocs/index.php/AAAI/AAAI18/paper/view/17410} {Learning
  interpretable spatial operations in a rich 3d blocks world}.
\newblock In \emph{Proceedings of the Thirty-Second {AAAI} Conference on
  Artificial Intelligence, (AAAI-18), the 30th innovative Applications of
  Artificial Intelligence (IAAI-18), and the 8th {AAAI} Symposium on
  Educational Advances in Artificial Intelligence (EAAI-18), New Orleans,
  Louisiana, USA, February 2-7, 2018}, pages 5028--5036. {AAAI} Press.

\bibitem[{Bisk et~al.(2016{\natexlab{b}})Bisk, Yuret, and
  Marcu}]{bisk2016@blockGroundingModel}
Yonatan Bisk, Deniz Yuret, and Daniel Marcu. 2016{\natexlab{b}}.
\newblock \href {https://doi.org/10.18653/v1/N16-1089} {Natural language
  communication with robots}.
\newblock In \emph{Proceedings of the 2016 Conference of the North {A}merican
  Chapter of the Association for Computational Linguistics: Human Language
  Technologies}, pages 751--761, San Diego, California. Association for
  Computational Linguistics.

\bibitem[{Burgoon et~al.(2013)Burgoon, Henderson, and
  Markman}]{burgoon2013abstractionPsychology}
Erin~M. Burgoon, Marlone~D. Henderson, and Arthur~B. Markman. 2013.
\newblock There are many ways to see the forest for the trees: A tour guide for
  abstraction.
\newblock \emph{Perspectives on Psychological Science}, 8(5):501--520.

\bibitem[{Chen et~al.(2019)Chen, Suhr, Misra, Snavely, and
  Artzi}]{chen2019@touchdown}
Howard Chen, Alane Suhr, Dipendra Misra, Noah Snavely, and Yoav Artzi. 2019.
\newblock \href {https://doi.org/10.1109/CVPR.2019.01282} {Touchdown: Natural
  language navigation and spatial reasoning in visual street environments}.
\newblock In \emph{2019 IEEE/CVF Conference on Computer Vision and Pattern
  Recognition (CVPR)}, pages 12530--12539.

\bibitem[{Chevalier-Boisvert et~al.(2018)Chevalier-Boisvert, Bahdanau, Lahlou,
  Willems, Saharia, Nguyen, and Bengio}]{chevalier2018babyai}
Maxime Chevalier-Boisvert, Dzmitry Bahdanau, Salem Lahlou, Lucas Willems,
  Chitwan Saharia, Thien~Huu Nguyen, and Yoshua Bengio. 2018.
\newblock Baby{AI}: A platform to study the sample efficiency of grounded
  language learning.
\newblock In \emph{International Conference on Learning Representations}.

\bibitem[{Clark and
  Wilkes-Gibbs(1986)}]{clark&Wilkes-Gibbs1986@referringCollaborative}
Herbert~H. Clark and Deanna Wilkes-Gibbs. 1986.
\newblock Referring as a collaborative process.
\newblock \emph{Cognition}, 22(1):1--39.

\bibitem[{Cuny et~al.(2010)Cuny, Snyder, and
  Wing}]{cuny&wing2010@demystifyingCT}
Jan Cuny, Larry Snyder, and Jeannette~M. Wing. 2010.
\newblock \emph{Demystifying Computational Thinking for Non-computer
  Scientists}.
\newblock Unpublished manuscript in progress, referenced in
  \url{http://www.cs.cmu.edu/\~CompThink/resources/TheLinkWing.pdf}.

\bibitem[{Denning et~al.(1989)Denning, Comer, Gries, Mulder, Tucker, Turner,
  Young, and Denning}]{denning1989@computingDiscipline}
Peter~J. Denning, Douglas~E. Comer, David Gries, Michael~C. Mulder, Allen
  Tucker, A.~Joe Turner, Paul~R. Young, and Peter~J. Denning. 1989.
\newblock \href {https://doi.org/10.1145/63238.63239} {Computing as a
  discipline}.
\newblock \emph{Communication of the ACM}, 32(1):9–23.

\bibitem[{Dijkstra(1972)}]{dijkstra1972@humbleProgrammer}
Edsger~W. Dijkstra. 1972.
\newblock The humble programmer.
\newblock \emph{ACM Turing Award Lectures}.

\bibitem[{Dijkstra(Accessed 1 May 2021)}]{dijkestra@abstractQuote2}
Edsger~W. Dijkstra. Accessed 1 May 2021.
\newblock \href
  {http://web.archive.org/web/20080207010024/http://www.808multimedia.com/winnt/kernel.htm}
  {Ew dijkstra quotes}.

\bibitem[{Finegan-Dollak et~al.(2018)Finegan-Dollak, Kummerfeld, Zhang,
  Ramanathan, Sadasivam, Zhang, and Radev}]{finegan-dollak2018@text2SQL}
Catherine Finegan-Dollak, Jonathan~K. Kummerfeld, Li~Zhang, Karthik Ramanathan,
  Sesh Sadasivam, Rui Zhang, and Dragomir Radev. 2018.
\newblock \href {https://doi.org/10.18653/v1/P18-1033} {Improving text-to-{SQL}
  evaluation methodology}.
\newblock In \emph{Proceedings of the 56th Annual Meeting of the Association
  for Computational Linguistics (Volume 1: Long Papers)}, pages 351--360,
  Melbourne, Australia. Association for Computational Linguistics.

\bibitem[{FitzGerald et~al.(2018)FitzGerald, Michael, He, and
  Zettlemoyer}]{fitzgerald2018@LargeScaleQASRL}
Nicholas FitzGerald, Julian Michael, Luheng He, and Luke Zettlemoyer. 2018.
\newblock \href {https://doi.org/10.18653/v1/P18-1191} {Large-scale {QA}-{SRL}
  parsing}.
\newblock In \emph{Proceedings of the 56th Annual Meeting of the Association
  for Computational Linguistics (Volume 1: Long Papers)}, pages 2051--2060,
  Melbourne, Australia. Association for Computational Linguistics.

\bibitem[{Ginat and Blau(2017)}]{Ginat2017@multipleAbstraction1}
David Ginat and Yoav Blau. 2017.
\newblock \href {https://doi.org/10.1145/3017680.3017801} {Multiple levels of
  abstraction in algorithmic problem solving}.
\newblock In \emph{Proceedings of the 2017 ACM SIGCSE Technical Symposium on
  Computer Science Education}, SIGCSE '17, page 237–242, New York, NY, USA.
  Association for Computing Machinery.

\bibitem[{Goldman et~al.(2022)Goldman, Guriel, and Tsarfaty}]{goldman22lemma}
Omer Goldman, David Guriel, and Reut Tsarfaty. 2022.
\newblock \href {https://doi.org/10.18653/v1/2022.acl-short.96} {(un)solving
  morphological inflection: Lemma overlap artificially inflates models{'}
  performance}.
\newblock In \emph{Proceedings of the 60th Annual Meeting of the Association
  for Computational Linguistics (Volume 2: Short Papers)}, pages 864--870,
  Dublin, Ireland. Association for Computational Linguistics.

\bibitem[{Grice(1975)}]{grice1975@logic}
Herbert~P. Grice. 1975.
\newblock Logic and conversation.
\newblock In \emph{Speech Acts}, pages 41--58. Brill.

\bibitem[{Grover and Pea(2013)}]{grover&pea2013@CT}
Shuchi Grover and Roy Pea. 2013.
\newblock Computational thinking in k--12: A review of the state of the field.
\newblock \emph{Educational researcher}, 42(1):38--43.

\bibitem[{Gwet(2015)}]{gwet2015@krippendorffAlpha}
Kilem~L. Gwet. 2015.
\newblock \href
  {https://www.researchgate.net/profile/Kilem-Gwet/publication/267823285_On_Krippendorff's_Alpha_Coefficient/links/60e3bf0892851ca944ae25d6/On-Krippendorffs-Alpha-Coefficient.pdf}
  {On {K}rippendorff’s alpha coefficient}.
\newblock Accessed: 1 June 2022.

\bibitem[{Haber et~al.(2019)Haber, Baumg{\"a}rtner, Takmaz, Gelderloos, Bruni,
  and Fern{\'a}ndez}]{haber2019@photobook}
Janosch Haber, Tim Baumg{\"a}rtner, Ece Takmaz, Lieke Gelderloos, Elia Bruni,
  and Raquel Fern{\'a}ndez. 2019.
\newblock \href {https://doi.org/10.18653/v1/P19-1184} {The {P}hoto{B}ook
  dataset: Building common ground through visually-grounded dialogue}.
\newblock In \emph{Proceedings of the 57th Annual Meeting of the Association
  for Computational Linguistics}, pages 1895--1910, Florence, Italy.
  Association for Computational Linguistics.

\bibitem[{Haberman(2004)}]{haberman2004@proceduralAbstraction}
Bruria Haberman. 2004.
\newblock High-school students' attitudes regarding procedural abstraction.
\newblock \emph{Education and Information Technologies}, 9(2):131--145.

\bibitem[{He et~al.(2020)He, Liu, Gao, and Chen}]{he2020deberta}
Pengcheng He, Xiaodong Liu, Jianfeng Gao, and Weizhu Chen. 2020.
\newblock Deberta: Decoding-enhanced bert with disentangled attention.
\newblock In \emph{International Conference on Learning Representations}.

\bibitem[{Herzig and Berant(2020)}]{herzig2020span}
Jonathan Herzig and Jonathan Berant. 2020.
\newblock Span-based semantic parsing for compositional generalization.
\newblock \emph{arXiv preprint arXiv:2009.06040. Version 2.}

\bibitem[{Jayannavar et~al.(2020)Jayannavar, Narayan-Chen, and
  Hockenmaier}]{jayannavar2020@minecraftBuilder}
Prashant Jayannavar, Anjali Narayan-Chen, and Julia Hockenmaier. 2020.
\newblock \href {https://www.aclweb.org/anthology/2020.acl-main.232} {Learning
  to execute instructions in a {M}inecraft dialogue}.
\newblock In \emph{Proceedings of the 58th Annual Meeting of the Association
  for Computational Linguistics}, pages 2589--2602, Online. Association for
  Computational Linguistics.

\bibitem[{Khani et~al.(2018)Khani, Goodman, and
  Liang}]{khaniGoodmanLiang2018@InfoJigsaw}
Fereshte Khani, Noah Goodman, and Percy Liang. 2018.
\newblock Planning, inference, and pragmatics in sequential language games.
\newblock \emph{Transactions of the Association for Computational Linguistics},
  6:543--555.

\bibitem[{Kim et~al.(2019)Kim, Kitaev, Chen, Rohrbach, Zhang, Tian, Batra, and
  Parikh}]{kim2019@codraw}
Jin-Hwa Kim, Nikita Kitaev, Xinlei Chen, Marcus Rohrbach, Byoung-Tak Zhang,
  Yuandong Tian, Dhruv Batra, and Devi Parikh. 2019.
\newblock \href {https://doi.org/10.18653/v1/P19-1651} {{C}o{D}raw:
  Collaborative drawing as a testbed for grounded goal-driven communication}.
\newblock In \emph{Proceedings of the 57th Annual Meeting of the Association
  for Computational Linguistics}, pages 6495--6513, Florence, Italy.
  Association for Computational Linguistics.

\bibitem[{Koppelman and Van~Dijk(2010)}]{koppelman2010@teachingAbstraction}
Herman Koppelman and Betsy Van~Dijk. 2010.
\newblock Teaching abstraction in introductory courses.
\newblock In \emph{Proceedings of the fifteenth annual conference on Innovation
  and technology in computer science education}, pages 174--178.

\bibitem[{Krippendorff(2004)}]{Krippendorff2004@krippendorffAlpha}
Klaus Krippendorff. 2004.
\newblock \emph{Content Analysis: An Introduction to Its Methodology (2nd ed.),
  Chapter 11}.
\newblock Beverly Hills, CA: Sage.

\bibitem[{Liu et~al.(2021)Liu, Yuan, Fu, Jiang, Hayashi, and
  Neubig}]{liuNeubig2021@NNPrompting}
Pengfei Liu, Weizhe Yuan, Jinlan Fu, Zhengbao Jiang, Hiroaki Hayashi, and
  Graham Neubig. 2021.
\newblock Pre-train, prompt, and predict: {A} systematic survey of prompting
  methods in natural language processing.
\newblock \emph{arXiv preprint arXiv:2107.13586. Version 1.}

\bibitem[{Long et~al.(2016)Long, Pasupat, and Liang}]{long2016@scone}
Reginald Long, Panupong Pasupat, and Percy Liang. 2016.
\newblock \href {https://doi.org/10.18653/v1/P16-1138} {Simpler
  context-dependent logical forms via model projections}.
\newblock In \emph{Proceedings of the 54th Annual Meeting of the Association
  for Computational Linguistics (Volume 1: Long Papers)}, pages 1456--1465,
  Berlin, Germany. Association for Computational Linguistics.

\bibitem[{MacMahon et~al.(2006)MacMahon, Stankiewicz, and
  Kuipers}]{MacMahon2006@sail}
Matt MacMahon, Brian Stankiewicz, and Benjamin Kuipers. 2006.
\newblock Walk the talk: Connecting language, knowledge, and action in route
  instructions.
\newblock In \emph{Proceedings of the 21st National Conference on Artificial
  Intelligence - Volume 2}, AAAI'06, page 1475–1482. AAAI Press.

\bibitem[{Misra et~al.(2018)Misra, Bennett, Blukis, Niklasson, Shatkhin, and
  Artzi}]{misraArtzi2018@navigationLaniChai}
Dipendra Misra, Andrew Bennett, Valts Blukis, Eyvind Niklasson, Max Shatkhin,
  and Yoav Artzi. 2018.
\newblock \href {https://doi.org/10.18653/v1/D18-1287} {Mapping instructions to
  actions in 3{D} environments with visual goal prediction}.
\newblock In \emph{Proceedings of the 2018 Conference on Empirical Methods in
  Natural Language Processing}, pages 2667--2678, Brussels, Belgium.
  Association for Computational Linguistics.

\bibitem[{Paz-Argaman and Tsarfaty(2019)}]{tzuf2019@RUN}
Tzuf Paz-Argaman and Reut Tsarfaty. 2019.
\newblock \href {https://www.aclweb.org/anthology/D19-1681} {{RUN} through the
  streets: A new dataset and baseline models for realistic urban navigation}.
\newblock In \emph{Proceedings of the 2019 Conference on Empirical Methods in
  Natural Language Processing and the 9th International Joint Conference on
  Natural Language Processing (EMNLP-IJCNLP)}, pages 6450--6456, Hong Kong,
  China. Association for Computational Linguistics.

\bibitem[{Pi{\v{s}}l and Mare{\v{c}}ek(2017)}]{pisl&marecek2017@BlocksRNN}
Bed{\v{r}}ich Pi{\v{s}}l and David Mare{\v{c}}ek. 2017.
\newblock Communication with robots using multilayer recurrent networks.
\newblock In \emph{Proceedings of the First Workshop on Language Grounding for
  Robotics}, pages 44--48.

\bibitem[{Pyatkin et~al.(2020)Pyatkin, Klein, Tsarfaty, and
  Dagan}]{pyatkin2020@QAdiscourse}
Valentina Pyatkin, Ayal Klein, Reut Tsarfaty, and Ido Dagan. 2020.
\newblock \href {https://doi.org/10.18653/v1/2020.emnlp-main.224}
  {{QAD}iscourse - discourse relations as {QA} pairs: {R}epresentation,
  crowdsourcing and baselines}.
\newblock In \emph{Proceedings of the 2020 Conference on Empirical Methods in
  Natural Language Processing (EMNLP)}, pages 2804--2819, Online. Association
  for Computational Linguistics.

\bibitem[{Raffel et~al.(2020)Raffel, Shazeer, Roberts, Lee, Narang, Matena,
  Zhou, Li, and Liu}]{2020t5}
Colin Raffel, Noam Shazeer, Adam Roberts, Katherine Lee, Sharan Narang, Michael
  Matena, Yanqi Zhou, Wei Li, and Peter~J. Liu. 2020.
\newblock \href {http://jmlr.org/papers/v21/20-074.html} {Exploring the limits
  of transfer learning with a unified text-to-text transformer}.
\newblock \emph{Journal of Machine Learning Research}, 21(140):1--67.

\bibitem[{Relkin and Bers(2019)}]{relkin2019@assessmentCT}
Emily Relkin and Marina~Umaschi Bers. 2019.
\newblock Designing an assessment of computational thinking abilities for young
  children.
\newblock In Lynn~E. Cohen and Sandra Waite-Stupiansky, editors, \emph{STEM in
  Early Childhood Education: How Science, Technology, Engineering, and
  Mathematics Strengthen Learning}, chapter~5. Routledge, New York.

\bibitem[{Roit et~al.(2020)Roit, Klein, Stepanov, Mamou, Michael, Stanovsky,
  Zettlemoyer, and Dagan}]{roit2020@controlledCrowdsource}
Paul Roit, Ayal Klein, Daniela Stepanov, Jonathan Mamou, Julian Michael,
  Gabriel Stanovsky, Luke Zettlemoyer, and Ido Dagan. 2020.
\newblock \href {https://doi.org/10.18653/v1/2020.acl-main.626} {Controlled
  crowdsourcing for high-quality {QA}-{SRL} annotation}.
\newblock In \emph{Proceedings of the 58th Annual Meeting of the Association
  for Computational Linguistics}, pages 7008--7013, Online. Association for
  Computational Linguistics.

\bibitem[{Ructtinger and Stevens(2017)}]{Ructtinger2017@taskDesignCT}
Liliana Ructtinger and Robert Stevens. 2017.
\newblock \href
  {https://search.informit.org/doi/10.3316/informit.367136823707035}
  {Computational thinking task design and assessment.}
\newblock \emph{Scan: The Journal For Educators}, 36(1):34–41.

\bibitem[{Shute et~al.(2017)Shute, Sun, and
  Asbell-Clarke}]{shute2017demystifyingCT}
Valerie~J. Shute, Chen Sun, and Jodi Asbell-Clarke. 2017.
\newblock Demystifying computational thinking.
\newblock \emph{Educational Research Review}, 22:142--158.

\bibitem[{Suhr et~al.(2019)Suhr, Yan, Schluger, Yu, Khader, Mouallem, Zhang,
  and Artzi}]{suhr2019@CerealBar}
Alane Suhr, Claudia Yan, Jack Schluger, Stanley Yu, Hadi Khader, Marwa
  Mouallem, Iris Zhang, and Yoav Artzi. 2019.
\newblock \href {https://doi.org/10.18653/v1/D19-1218} {Executing instructions
  in situated collaborative interactions}.
\newblock In \emph{Proceedings of the 2019 Conference on Empirical Methods in
  Natural Language Processing and the 9th International Joint Conference on
  Natural Language Processing (EMNLP-IJCNLP)}, pages 2119--2130, Hong Kong,
  China. Association for Computational Linguistics.

\bibitem[{Udagawa and Aizawa(2019)}]{Takuma&Akiko2019@oneCommon}
Takuma Udagawa and Akiko Aizawa. 2019.
\newblock \href {https://doi.org/10.1609/aaai.v33i01.33017120} {A natural
  language corpus of common grounding under continuous and partially-observable
  context}.
\newblock \emph{Proceedings of the AAAI Conference on Artificial Intelligence},
  33(01):7120--7127.

\bibitem[{Wang et~al.(2017)Wang, Ginn, Liang, and
  Manning}]{wang2017@naturalizingVoxelurn}
Sida~I. Wang, Samuel Ginn, Percy Liang, and Christopher~D. Manning. 2017.
\newblock \href {https://doi.org/10.18653/v1/P17-1086} {Naturalizing a
  programming language via interactive learning}.
\newblock In \emph{Proceedings of the 55th Annual Meeting of the Association
  for Computational Linguistics (Volume 1: Long Papers)}, pages 929--938,
  Vancouver, Canada. Association for Computational Linguistics.

\bibitem[{Wang et~al.(2016)Wang, Liang, and
  Manning}]{wang2016@SHRDLURNWittgensteinLanguageGame}
Sida~I. Wang, Percy Liang, and Christopher~D. Manning. 2016.
\newblock \href {https://doi.org/10.18653/v1/P16-1224} {Learning language games
  through interaction}.
\newblock In \emph{Proceedings of the 54th Annual Meeting of the Association
  for Computational Linguistics (Volume 1: Long Papers)}, pages 2368--2378,
  Berlin, Germany. Association for Computational Linguistics.

\bibitem[{Wing(2017)}]{wing2017@CTinfluence}
Jeannette Wing. 2017.
\newblock Computational thinking’s influence on research and education for
  all.
\newblock \emph{Italian Journal of Educational Technology}, 25(2):7--14.

\bibitem[{Wing(2011)}]{wing2011@CT}
Jeannette~M. Wing. 2011.
\newblock Computational thinking—what and why?
\newblock \textit{CMU Research Notebook}.

\bibitem[{Wolf et~al.(2020)Wolf, Debut, Sanh, Chaumond, Delangue, Moi, Cistac,
  Rault, Louf, Funtowicz, Davison, Shleifer, von Platen, Ma, Jernite, Plu, Xu,
  Le~Scao, Gugger, Drame, Lhoest, and Rush}]{wolf2020transformers}
Thomas Wolf, Lysandre Debut, Victor Sanh, Julien Chaumond, Clement Delangue,
  Anthony Moi, Pierric Cistac, Tim Rault, R{\'e}mi Louf, Morgan Funtowicz, Joe
  Davison, Sam Shleifer, Patrick von Platen, Clara Ma, Yacine Jernite, Julien
  Plu, Canwen Xu, Teven Le~Scao, Sylvain Gugger, Mariama Drame, Quentin Lhoest,
  and Alexander Rush. 2020.
\newblock \href {https://doi.org/10.18653/v1/2020.emnlp-demos.6} {Transformers:
  State-of-the-art natural language processing}.
\newblock In \emph{Proceedings of the 2020 Conference on Empirical Methods in
  Natural Language Processing: System Demonstrations}, pages 38--45, Online.
  Association for Computational Linguistics.

\bibitem[{Xu et~al.(2022)Xu, Szlam, and Weston}]{xu2022@longTermDialogues}
Jing Xu, Arthur Szlam, and Jason Weston. 2022.
\newblock \href {https://doi.org/10.18653/v1/2022.acl-long.356} {Beyond
  goldfish memory: Long-term open-domain conversation}.
\newblock In \emph{Proceedings of the 60th Annual Meeting of the Association
  for Computational Linguistics (Volume 1: Long Papers)}, pages 5180--5197,
  Dublin, Ireland. Association for Computational Linguistics.

\end{thebibliography}
